\documentclass[journal,twoside,web]{ieeecolor}
\usepackage{tmi}
\usepackage{cite}
\usepackage{amsmath,amssymb,amsfonts}
\usepackage{algorithmic}
\usepackage{graphicx}
\usepackage{textcomp}
\usepackage{xcolor}
\usepackage{url}
\usepackage{booktabs}

\def\Fig#1{{Fig.~\ref{fig:#1}}}
\def\Eq#1{{Eq.~\ref{eq:#1}}}
\def\Table#1{{Table~\ref{tbl:#1}}}
\def\eg{{e.g.}}
\def\etal{{et al.}}
\def\ie{{i.e.}}

\def\BibTeX{{\rm B\kern-.05em{\sc i\kern-.025em b}\kern-.08em
		T\kern-.1667em\lower.7ex\hbox{E}\kern-.125emX}}
\begin{document}
	
	\title{Learning from Multiple Datasets with Heterogeneous and Partial Labels \\ for Universal Lesion Detection in CT}

	\author{Ke Yan, Jinzheng Cai, \textit{Member, IEEE}, Youjing Zheng, Adam P.~Harrison, \textit{Member, IEEE}, Dakai Jin, \\Youbao Tang, Yuxing Tang, Lingyun Huang, Jing Xiao, Le Lu, \textit{Fellow, IEEE}
		\thanks{K.~Yan, J.~Cai, A.~P.~Harrison, D.~Jin, Y.~Tang, Y.~Tang, and L.~Lu are with PAII Inc., Bethesda, MD 20817, USA. (email: 
			yankethu@gmail.com, caijinzhengcn@gmail.com, adam.p.harrison@gmail.com, dakai.jin@gmail.com, tybxiaobao@gmail.com, tangyuxing87@gmail.com, tiger.lelu@gmail.com).}
		\thanks{Y. Zheng is with Virginia Polytechnic Institute and State University, Blacksburg, VA 24061, USA. (email: zhengyoujing@vt.edu)}
		\thanks{L. Huang, J. Xiao is with Ping An Insurance (Group) Company of China, Ltd., Shenzhen, 510852, PRC. (email: \{huanglingyun691, xiaojing661\}@pingan.com.cn)}
		\thanks{Corresponding author: Ke Yan}
}
	\maketitle
	
	\begin{abstract}
		Large-scale datasets with high-quality labels are desired for training accurate deep learning models. However, due to the annotation cost, datasets in medical imaging are often either partially-labeled or small. For example, DeepLesion is such a large-scale CT image dataset with lesions of various types, but it also has many unlabeled lesions (missing annotations). When training a lesion detector on a partially-labeled dataset, the missing annotations will generate incorrect negative signals and degrade the performance. Besides DeepLesion, there are several small single-type datasets, such as LUNA for lung nodules and LiTS for liver tumors. These datasets have heterogeneous label scopes, \ie, different lesion types are labeled in different datasets with other types ignored. In this work, we aim to develop a universal lesion detection algorithm to detect a variety of lesions. The problem of heterogeneous and partial labels is tackled. First, we build a simple yet effective lesion detection framework named Lesion ENSemble (LENS). LENS can efficiently learn from multiple heterogeneous lesion datasets in a multi-task fashion and leverage their synergy by proposal fusion. Next, we propose strategies to mine missing annotations from partially-labeled datasets by exploiting clinical prior knowledge and cross-dataset knowledge transfer. Finally, we train our framework on four public lesion datasets and evaluate it on 800 manually-labeled sub-volumes in DeepLesion. Our method brings a relative improvement of 49\% compared to the current state-of-the-art approach in the metric of average sensitivity. We have publicly released our manual 3D annotations of DeepLesion online.\footnote{\newline \url{https://github.com/viggin/DeepLesion_manual_test_set}}
	\end{abstract}
	
	
	\section{Introduction}
	\label{sec:introduction}
	\IEEEPARstart{T}{raining} data plays a key role in data-driven deep learning algorithms for medical image analysis. Different from natural images, annotating medical images is highly tedious and demands extensive clinical expertise, making it difficult to acquire large-scale medical image datasets with high-quality labels. A possible solution is to train one model on multiple small datasets to integrate their knowledge~\cite{Tajbakhsh2020imperfect, Huang2019U2, Zhou2019organ, Luo2020imperfect, Cohen2020domain, Dmitriev2019seg, Gundel2019Location}. Compared to training a separate model on each dataset, joint training offers three advantages: first, the size of training data is expanded without further manual annotation; second, training and inference become more efficient, as multiple models are consolidated into a single model~\cite{Tajbakhsh2020imperfect}; and third, the combined training data covers a larger distribution (\eg, different datasets come from different hospitals), potentially increasing the generalizability of the trained model~\cite{Luo2020imperfect}. The main challenge of this strategy is that the label scope of different datasets is often different. For example, in chest X-ray disease classification, the label set is not identical across datasets~\cite{Gundel2019Location, Cohen2020domain, Luo2020imperfect}; in multi-organ segmentation, varying organs are labeled in different datasets~\cite{Huang2019U2, Zhou2019organ, Dmitriev2019seg}. Even if a label is shared between two datasets, its definition may vary due to different data collection and annotation criteria (concept shift)~\cite{Cohen2020domain}. Therefore, combining multiple datasets is not straightforward and may degrade accuracy if these problems are not solved~\cite{Huang2019U2}.
	
	Another method to address data scarcity is to collect images and labels by data mining~\cite{Wang2017ChestXray, Yan2018DeepLesion}. It can produce large-scale datasets with small manual efforts, but the mined labels are often imperfect. Take the DeepLesion dataset~\cite{Yan2018graph, Yan2018DeepLesion} as an example. It was collected by mining lesion annotations directly from the picture archiving and communication system (PACS), which stores the lesion markers~\cite{Eisenhauer2009RECIST} already annotated by radiologists during their routine work. DeepLesion includes over 32K lesions on various body parts in computed tomography (CT) scans. 
	Along with its large scale and ease of collection, DeepLesion also has a limitation: not all lesions in every slice were annotated. This is because radiologists generally mark only representative lesions in each scan~\cite{Eisenhauer2009RECIST} in their routine work. This missing annotation or partial label problem will cause incorrect training signals (some negative proposals are actually positive), resulting in a lower detection accuracy.
	
	In this paper, we tackle the heterogeneous and partial label problem to aid large-scale multi-source deep learning in the lesion detection task. As a major task in medical image analysis, lesion detection aims to help radiologists locate abnormal image findings, so as to decrease reading time and improve accuracy~\cite{Litjens2017survey, Sahiner2018survey}. Existing lesion detection works commonly focus on lesions of specific types and organs. For example, lung nodules~\cite{Setio2017LUNA, Zhu2018DeepEM, dou2017multilevel}, liver tumors~\cite{Wang2019LiTS}, and lymph nodes~\cite{Roth2016randView, Shin2016TMICNN,zhu2020detecting} have been extensively studied. However, in clinical scenarios, a CT scan may contain multiple types of lesions in different organs. For instance, metastasis (\eg, lung cancer) can spread to regional lymph nodes and other body parts (\eg, liver, bone, adrenal, etc.). Clinicians need to locate all types of findings to determine the patient's tumor stage and future treatment~\cite{Eisenhauer2009RECIST}. The RSNA standard radiology report templates~\cite{RSNA2020template} also direct radiologists to examine multiple body parts. For chest CT, the RSNA template asks for findings in lungs, airways, pleural space, heart, pericardium, mediastinum, hila, thoracic vessels, bones, and chest wall. In order to meet this clinical need, universal lesion detection (ULD) is attracting increasing attention~\cite{Yan20183DCE, Wang2019universal, Li2019MVP, Wang2019LiTS, Yan2019MULAN}. ULD tries to find various lesions discoverable on a single CT scan, imitating what many radiologists commonly do in their daily work. It is more efficient and scalable than designing a special model for every lesion type. It can complement single-type models by finding relatively rare but still clinically significant lesion types that cannot be covered by single-type lesion detection models.
	
	\begin{figure}[t]
		\centerline{\includegraphics[width=\columnwidth,trim=110 100 270 100, clip]{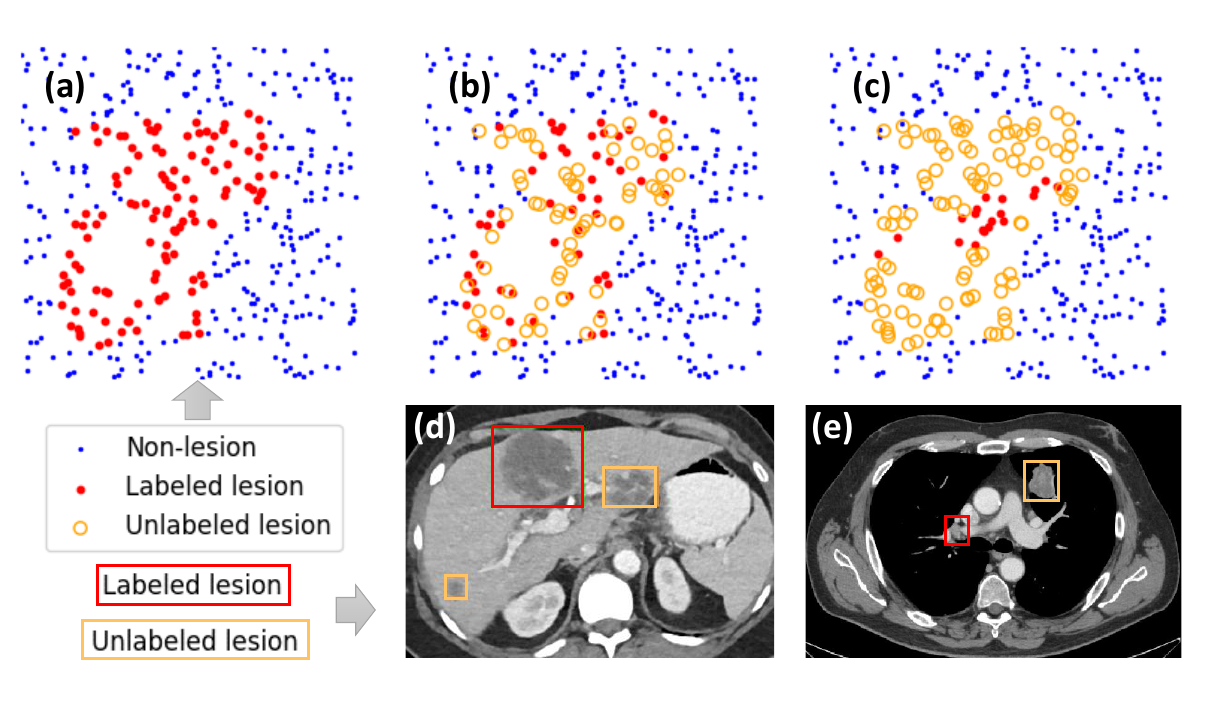}} 
		\caption{Illustration of the heterogeneous and partial label problem in lesion datasets. \textbf{(a)}. Simulated scatter map of a perfectly-labeled dataset. \textbf{(b)}. A partially-labeled universal dataset where a portion of lesions are labeled that cover various lesion types. \textbf{(d)} shows an example from the DeepLesion dataset~\cite{Yan2018DeepLesion}, where a liver lesion is labeled in the image but two smaller ones are not (missing annotations). \textbf{(c)}. A fully-labeled single-type dataset where all lesions of a certain type are labeled. \textbf{(e)} shows an example from the NIH lymph node dataset~\cite{NIH_LN_dataset}, where an enlarged mediastinal lymph node is labeled but a neighboring lung lesion is not.}
		\label{fig:scatter}
	\end{figure}
	
	Our first goal is to improve ULD by learning from multiple lesion datasets. 
	Existing works~\cite{Yan20183DCE, Wang2019universal, Li2019MVP, Wang2019LiTS, Yan2019MULAN} on ULD learned from the DeepLesion dataset alone. In this work, we make one step forward to also leverage other public single-type lesion datasets~\cite{Setio2017LUNA, Bilic2019LiTS, NIH_LN_dataset}, which provide annotations of specific lesion types. \Fig{scatter} shows exemplar labels in different lesion datasets. To deal with the label scope difference and concept shift mentioned above, we propose a simple yet effective multi-dataset lesion detection framework. It includes multiple dataset-specific anchor-free proposal networks and a multi-task detection head. Given an image, the algorithm can predict several groups of lesion proposals that match the semantics of each dataset. It can be considered as an ensemble of ``dataset experts'', thus is named Lesion ENSemble (LENS). 
	A patch-based 3D classifier is then used to further reduce false positives.
	
	We first train LENS on all datasets to generate lesion proposals on the training set of DeepLesion, and then mine missing annotations from them to mitigate the partial label problem. 
	We use cross-slice box propagation to extend 2D annotations to 3D. Next, we propose an intra-patient lesion matching strategy to mine lesions that are annotated in one scan but missed in another scan of the same patient, leveraging the prior knowledge that the same lesion instance exists across scans of the same patient. An embedding-based retrieval method is adopted for matching. Lastly, we propose a cross-dataset lesion mining strategy to find more uncertain lesions with the help of single-type dataset experts in LENS. The mined missing annotations and uncertain lesions are incorporated to retrain LENS for performance improvement. The strategies also enable us to mine lesions from the abundant unlabeled images in DeepLesion and then leverage them during training.
	
	In our framework, knowledge in multiple datasets are integrated in three levels: 1) different datasets share the network backbone of LENS to learn better feature representation from multi-source CT images; 2) the lesion proposals of multiple dataset experts in LENS are fused to improve the sensitivity of ULD; 3) single-type datasets help to mine missing annotations in partially-labeled datasets to improve the quality of training labels. We employ DeepLesion as well as three single-type datasets in our framework, namely LUNA (LUng Nodule Analysis)~\cite{Setio2017LUNA}, LiTS (Liver Tumor Segmentation Benchmark)~\cite{Bilic2019LiTS}, and NIH-LN (NIH Lymph Node)~\cite{NIH_LN_dataset}. For evaluation, we manually annotated all lesions in 800 sub-volumes in DeepLesion as the test set\footnote{\scriptsize We were unable to annotate full volumes, since images in DeepLesion were released in sub-volumes containing 7$ \sim $220 consecutive slices.}. On this challenging task, our LENS trained on multiple datasets outperforms the current single-dataset state-of-the-art method~\cite{Yan2019MULAN} in average sensitivity (from 33.9\% to 39.4\%). After adding the mined lesions, the sensitivity is further improved to 47.6\%. It is not our goal to achieve new state-of-the-art results on the single-type lesion datasets. Nevertheless, we found that LENS jointly trained on 4 datasets achieved comparable or better accuracy on each dataset relative to the baselines trained on each dataset alone, and significantly outperformed the baselines when the number of training images is small.
	
	The main contributions of this paper are summarized as follows: 1) The large-scale heterogeneous dataset fusion problem in lesion detection is tackled for the first time via our LENS network. 2) We propose two novel strategies, \ie~intra-patient lesion matching and cross-dataset lesion mining, to alleviate the missing annotation problem and improve lesion detection performance. 3) Knowledge is integrated across datasets through feature sharing, proposal fusion, and annotation mining. 4) The ULD accuracy on DeepLesion~\cite{Yan2018DeepLesion} is significantly improved upon previous state-of-the-art work~\cite{Yan2019MULAN}.
	
	\section{Related Work}
	
	{\bf Universal lesion detection:} Convolutional neural networks (CNNs), such as Faster R-CNN~\cite{Ren2015faster} and Mask R-CNN~\cite{He2017MaskRCNN}, are widely-used in lesion detection. Based on these detection networks, ULD has been improved by researchers using 3D context fusion~\cite{Yan20183DCE, Yan2019MULAN, Wang2019LiTS}, attention mechanisms~\cite{Wang2019universal, Li2019MVP, Wang2019LiTS}, multi-task learning~\cite{Yan2019MULAN, Wang2019universal}, and hard negative mining~\cite{Tang2019Uldor}. 3D context information in neighboring slices is important for detection, as lesions may be less distinguishable in just one 2D axial slice. Volumetric attention~\cite{Wang2019LiTS} exploited 3D information with multi-slice image inputs and a 2.5D network and obtained top results on the LiTS dataset. In~\cite{Li2019MVP, Wang2019LiTS}, attention mechanisms were applied to emphasize important regions and channels in feature maps. The multi-task universal lesion analysis network (MULAN)~\cite{Yan2019MULAN} achieved the state-of-the-art accuracy on DeepLesion with a 3D feature fusion strategy and joint learning of lesion detection, segmentation, and tagging tasks. However, it did not handle the missing annotations. ULDor~\cite{Tang2019Uldor} used a trained detector to mine hard negative proposals and then retrained the model, but the mined negatives may actually contain positives because of missing annotations. None of the above methods can deal with multiple datasets with heterogeneous labels.
	
	{\bf Multi-task and multi-dataset learning:} 
	To increase training data, reduce overfitting, and improve accuracy, it is sometimes required to learn from multiple datasets labeled by different institutes using varying criteria~\cite{Tajbakhsh2020imperfect}. In chest X-ray classification, it is found that joint training on multiple datasets leads to better performance~\cite{Gundel2019Location, Lenga2020Continual}. Cohen \etal~\cite{Cohen2020domain} observed that the same class label had different distribution (concept shift) between multiple chest X-ray datasets and simply pooling all datasets is not optimal. Luo \etal~\cite{Luo2020imperfect} applied model ensemble to mine missing labels in heterogeneous datasets. In multi-organ segmentation, Zhou \etal~\cite{Zhou2019organ} and Dmitriev \etal~\cite{Dmitriev2019seg} studied how to learn multi-organ segmentation from single-organ datasets, incorporating priors on organ sizes and dataset-conditioned features, respectively. Although multi-dataset learning is receiving increasing attention in classification and segmentation tasks in medical imaging, it has not been sufficiently studied in lesion detection. The domain-attentive universal detector~\cite{Wang2019universal} used a domain attention module to learn DeepLesion as well as 10 other natural object detection datasets. Yet, it did not exploit the semantic overlap between datasets. Our framework leverages the synergy of lesion datasets both to learn shared features and to use their semantic overlaps for proposal fusion and annotation mining.
	
	{\bf Learning with partial labels:} In detection, knowledge distillation~\cite{Hinton2014distill} can help to find missing annotations. The basic idea is to treat the predicted boxes of a model as ground-truths. Predictions from multiple transformations of unlabeled data were merged to generate new training annotations in~\cite{Radosavovic2017data}. 
	Prior knowledge can also help to infer reliable missing annotations. Niitani \etal~\cite{Niitani2019sample} introduced part-aware sampling that assumes an object (car) must contain its parts (tire). Jin \etal~\cite{Jin2018mining} mined hard negative and positive proposals from unlabeled videos based on the prior that object proposals should be continuous across frames. Cai \etal~\cite{Cai2020harvest} iteratively mined missing annotations from DeepLesion with active learning by asking doctors to annotate a small portion of images. Wang \etal~\cite{Wang2019missing} propagated 2D annotations to adjacent slices to mine missing annotations in DeepLesion. In our framework, besides applying a similar strategy as \cite{Wang2019missing}, we also leverage prior knowledge of intra-patient lesion correspondence and cross-dataset knowledge distillation to find reliable and uncertain missing annotations.
	
	{\bf Domain adaptation and semi-supervised learning:} Domain adaptation is a type of transfer learning in which the source and target domains have similar tasks but different distributions~\cite{Pan2010transfer,Zhou2020survey}. In medical imaging, it has been extensively studied in cross-modality image segmentation with the help of image synthesis techniques~\cite{Tajbakhsh2020imperfect,Cai2019cycle,Liu2020JSSR}. These methods aim to align the images or features of multiple datasets with different modalities, so as to segment the same body structure. However, we aim to combine multiple datasets with the same modality but different tasks (varying lesion types annotated in each dataset), so our problem is different from domain adaptation. Semi-supervised learning~\cite{Tajbakhsh2020imperfect,Zhou2020survey} is another related research area in which not all data are labeled. In medical image segmentation, classification, and detection tasks, various methods have been proposed to exploit unlabeled data, such as co-training~\cite{Raju2020CoHeter,Xia2020semi,Zhou2019semi}, relation consistency regularization~\cite{Liu2020relation}, and MixUp augmentation~\cite{Wang2020focalmix}. Our work is slightly different from semi-supervised learning because all our data are labeled, but in an incomplete way.
	
	\begin{figure*}[h]
		\centerline{\includegraphics[width=.9\textwidth,trim=0 0 0 0, clip]{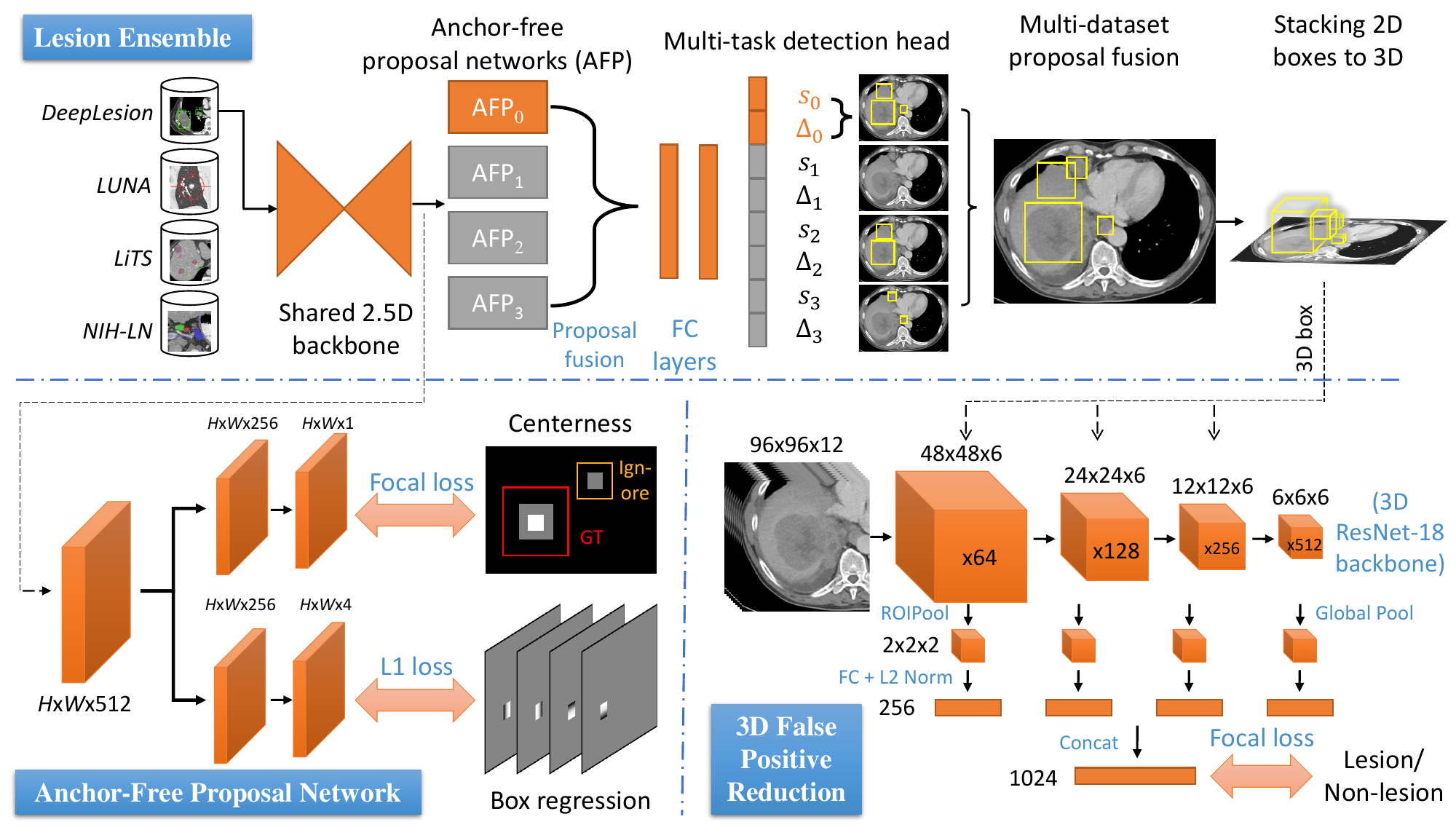}} 
		\caption{Framework of the proposed Lesion Ensemble (LENS), the anchor-free proposal network (AFP), and the 3D false positive reduction (FPR) network. AFP works as a part of LENS to generate initial proposals. FPR further classifies the 3D lesion proposals of LENS. LENS jointly learns from multiple datasets. When a training image is from one dataset (\eg~dataset 0), the parts in orange color are those to be updated.}
		\label{fig:LENS_framework}
	\end{figure*}
	
	\section{Method Overview}
	
	Our lesion detection framework combines multi-dataset learning with missing annotation mining. The former generates lesion proposals as the basis of the latter, while the latter provides refined labels to retrain the former. It consists of five main steps, which we will introduce in Sections \ref{sec:mtd:detector} and \ref{sec:mtd:mining}:
	
	\begin{enumerate}
		\item Train LENS on all datasets using existing annotations.
		\item Generate multi-dataset proposals on the training set of the partially-labeled datasets.
		\item Mine missing annotations and uncertain lesions from the generated proposals.
		\item Retrain LENS with the missing annotations as additional ground-truths and uncertain ones as ignored regions.
		\item Use the true positive and false positive proposals of LENS to train a classifier for false positive reduction.
	\end{enumerate}

	\section{Multi-dataset Lesion Detection}
	\label{sec:mtd:detector}
	
	The proposed detection framework is exhibited in \Fig{LENS_framework}. It consists of a two-stage detection network and a classification network for false positive reduction. The detection network, Lesion Ensemble (LENS), 
	contains a shared backbone, multiple simple yet effective anchor-free proposal networks, and a multi-task detection head.
	
	\subsection{Backbone}
	
	We make different datasets share the network backbone of LENS to learn better feature representation from multi-source CT images. Similar to \cite{Yan2019MULAN, Wang2019LiTS}, we use a 2.5D truncated DenseNet-121~\cite{Huang2017DenseNet} with 3D feature fusion layers and a feature pyramid network (FPN)~\cite{Lin2016Pyramid} as the backbone. 
	The detailed structure of the backbone can be found in~\cite{Yan2019MULAN}. We tried to apply domain adaptation layers~\cite{Rebuffi2018domain, Wang2019universal} in multi-domain learning literature, but no improvement in accuracy was found. This is probably because all datasets we used are CT images with small difference in the image domain, while the domain adaptation layers~\cite{Rebuffi2018domain, Wang2019universal} were designed for very different images.
	
	\subsection{Anchor-Free Proposal Network (AFP)}
	\label{subsec:mtd:AFP}
	
	Anchor-free detectors~\cite{Tian2019FCOS, Zhou2019CenterNet, Zhu2019FSAF} 
	do not need manually tuned anchor sizes~\cite{Ren2015faster}, thus are convenient particularly when multiple datasets have different size distributions. Our proposed anchor-free proposal network (AFP) is displayed in \Fig{LENS_framework}. 
	Inspired by~\cite{Tian2019FCOS, Zhou2019CenterNet, Zhu2019FSAF}, we use a centerness branch and a box regression branch to predict each pixel in the feature map. Both branches include two convolutional layers and a ReLU layer. The centerness branch tells whether a pixel is in the center region of a lesion. Denote the ground-truth box as $ B=(x,y,w,h) $ where $ (x,y) $ is its center and $ (w,h) $ its width and height. Then, we choose ratios $0<r_c<r_i<1$ to define $ B_\text{ctn}=(x,y,r_c w,r_c h) $ and $ B_\text{ign}=(x,y,r_i w,r_i h) $, thus $ B_\text{ctn} $ and $ B_\text{ign}-B_\text{ctn} $ are the lesion's center region and ignored region, respectively. In \Fig{LENS_framework}, the white and gray areas indicate the center region and the ignored region of a lesion box, respectively. We use $ r_c=0.2, r_i=0.5 $ in this paper~\cite{Zhu2019FSAF}. The centerness branch is required to predict 1 in $ B_\text{ctn} $ and 0 in everywhere else except the ignored region (which will be ignored in the loss function). If a lesion box is marked as uncertain (Sec.~\ref{subsec:mam:CrossMine}), it will only have $ B_\text{ign} $ and have no $ B_\text{ctn} $ so as to be entirely ignored. The focal loss~\cite{Lin2017focal, Zhou2019CenterNet} is adopted to supervise the centerness branch: 
	\begin{equation}
		L^{\text{center}}=-0.5(\sum_{i\in B_{\text{ctn}}}(1-p_i)^2\log p_i + \sum_{i\notin B_{\text{ign}}}p_i^2\log(1-p_i) ),
	\end{equation}
	where $ p_i $ is the centerness prediction of pixel $ i $. The box regression branch predicts 4 values for each pixel. It uses the L1 loss, which is only computed in $ B_\text{ctn} $ of each ground-truth:
	\begin{equation}
		L^{\text{size}} = \frac{1}{n}\sum_{i\in B_{\text{ctn}}}\sum_{j=1}^4|\hat v_{ij}-v_{ij}|,
	\end{equation}
	where $ n $ is the total number of pixels in $B_{\text{ctn}}$, $\hat v_{ij}$ and $v_{ij}$ are the true and predicted distances between pixel $ i $ and the top, bottom, left, and right borders of the lesion box, respectively. During inference, a box is predicted on each pixel according to the four regressed values with the objectness score predicted by the centerness branch. Then, non-maximum suppression (NMS) is performed to aggregate all predicted boxes.
	
	\subsection{Lesion Ensemble (LENS)}
	\label{subsec:mtd:LENS}
	
	We use a two-stage detection framework due to its high accuracy, meanwhile employ a combination of data-specific AFPs and a multi-task detection head. In our problem, multiple lesion datasets have heterogeneous labels. A lesion type may be annotated in dataset $ i $ but not in $ j $. Therefore, it is suitable to learn them in a multi-task fashion~\cite{Wang2019universal}. LENS includes $ d $ dataset-specific AFPs (\Fig{LENS_framework}), where $ d $ is the number of datasets. We pool the proposals from all AFPs, do non-maximum suppression (NMS), then fed them to an ROIAlign layer~\cite{He2017MaskRCNN} 
	and a detection head. The detection head includes two fully connected (FC) layers that are shared between datasets. Then, dataset-specific classification layers and box regression layers predict the detection score $ s $ (lesion vs.~non-lesion) and box offsets $ \Delta $~\cite{Ren2015faster} for each proposal. To sum up, we first merge the proposals of all AFPs to generate a comprehensive set of dataset-agnostic and universal lesion proposals, then use the multi-task detection head to give each proposal multiple detection scores matching each dataset's semantics. Note that the entire LENS network is trained in one training process. The DenseNet+FPN backbone and the two FC layers in the detection head are updated in each training iteration, whereas a dataset-specific AFP or detection layer is updated only when the training data come from the corresponding dataset. The overall loss function of LENS is
	\begin{equation}\label{eq:loss}
	L=\sum_{i=1}^{d} \sum_{j=1}^{n_i} L_{ij}^{\text{center}} + \lambda_1 L_{ij}^{\text{size}} + L_{ij}^{\text{class}} + \lambda_2 L_{ij}^{\text{box}},
	\end{equation}
	where $ n_i $ is the number of training samples in dataset $ i $; $ L_i^{\text{center}} $ and $ L_i^{\text{size}} $ are the losses for the centerness and box regression branches of AFP$_i$, respectively; $ L_i^{\text{class}} $ and $ L_i^{\text{box}} $ are the cross-entropy and smooth L1 \cite{Ren2015faster} losses for the $i$th classification and box regression layers of the detection head, respectively. $ \lambda_1 $ and $ \lambda_2 $ are the loss weights, which we empirically set as 0.1 and 10 in this paper. In inference, LENS can generate multiple groups of lesion proposals in the detection head to match the semantics of each dataset for each test image. It can be considered as an ensemble of ``dataset experts''. This is more efficient than training a separate model for each dataset and run every model during inference. Our experiments demonstrate that the the accuracy on each dataset is comparable or improved owing to joint training, especially for small datasets. 
	
	
	\begin{figure}[]
		\centerline{\includegraphics[width=.9\columnwidth,trim=0 0 450 0, clip]{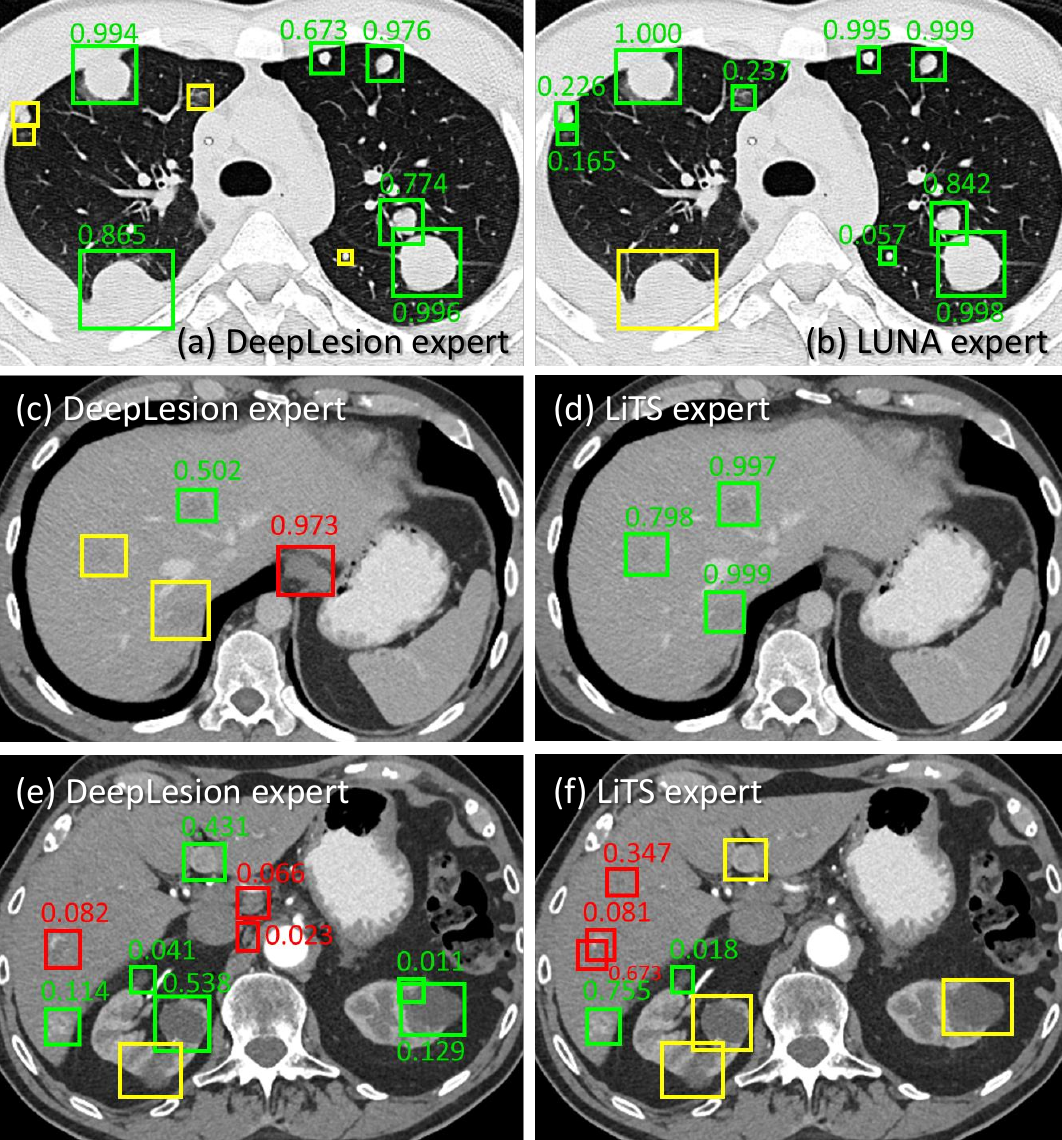}} 
		\caption{Predictions of different dataset experts of LENS. Green, red, and yellow boxes indicate true positives (TP), false positives (FP), and false negatives (FN), respectively. Predictions with detection scores smaller than 0.01 are not drawn. In \textbf{(a)}, the DeepLesion expert missed four small lung nodules but the LUNA expert in \textbf{(b)} gave them higher scores. However, the LUNA expert failed to detect a lung metastasis (in yellow box), while the DeepLesion expert found it. In \textbf{(c)}, the DeepLesion expert missed two indistinct liver lesions, but the LiTS expert in \textbf{(d)} detected all three with high confidence scores. There are also cases where the single-type expert is not helpful. In \textbf{(f)}, the LiTS expert generated more FPs in liver than the DeepLesion expert in \textbf{(e)}. It also failed to detect the kidney lesions and the liver metastasis in the left liver lobe.} 
		\label{fig:DL_spec_det_cmp}
	\end{figure}
	
	We find the predictions of dataset experts are complementary. As shown in \Fig{DL_spec_det_cmp}, the single-type experts often perform better in their specialties compared to the universal dataset expert. This is mainly because their training datasets are fully-labeled and include more hard cases. On the other hand, the single-type experts cannot detect other lesion types (\Fig{DL_spec_det_cmp} (f)). Even if the lesion is in the same organ of their specialty, they may still miss it because their datasets are limited to one lesion type, and lesions of certain appearances, sizes, or contrast phases may be uncommon in them (\Fig{DL_spec_det_cmp} (b)(f)). They may also generate more FPs if their datasets  
	have a different distribution (patient population, contrast phase, etc.) than the target dataset (\Fig{DL_spec_det_cmp} (f)). Therefore, a model trained on one single dataset may struggle to achieve the best performance in practical applications. 
	In medical diagnosis, human generalists and specialists can cooperate to combine their knowledge. We propose synergize the dataset experts by fusing their proposals to improve the final detection recall, since it is important for radiologists not to miss critical findings. For each test image, we generate proposals from all dataset experts, then do NMS to filter the overlapped boxes. The last step is to stack the predicted 2D boxes to 3D ones if the intersection over union (IoU) of two 2D boxes in consecutive slices is greater than $ \theta $. The $ x,y $ coordinates of the final 3D box is the average of the 2D ones weighted by their detection scores.

	\subsection{3D False Positive Reduction (FPR)}
	
	The predicted 3D boxes from LENS will undergo another round of false positive reduction (FPR), which has been proven effective in previous studies on lung nodule detection~\cite{Setio2017LUNA}. 
	The network is shown in \Fig{LENS_framework}. Its input is a 3D image patch whose center is the center of a 3D box. 
	We convert the $ 3\times 3 $ Conv filters of a ResNet-18~\cite{He2016resnet} to $ 1\times 3 \times 3 $ in ResBlocks 1--3 and to $ 3\times 3\times 3 $ in ResBlock 4~\cite{Yang2019acs}, which we found is better than converting all filters to $ 3\times 3\times 3 $. To encode multi-scale information, we use ROI pooling to crop feature maps from 4 ResBlocks. 
	The ROI of ResBlocks 1--3 is the 3D box and that of ResBlock4 is the whole 3D patch. Focal loss is adopted as the loss function. The final score of a lesion proposal is
	\begin{equation}\label{eq:final_lesion_score}
	s=(s_{\text{LENS}}+s_{\text{FPR}})/2,
	\end{equation}
	\ie, the average of the detection and classification networks. 
	FPR is helpful because it can focus on differentiating hard negative (HN) samples, namely the FPs from the detector. However, when the dataset is partially labeled, the HNs may actually be true lesions. Therefore, the missing annotations and uncertain lesions need to be removed from the FPs.
	
	\newpage
	\section{Missing Annotation Mining in Partially-Labeled Dataset}
	\label{sec:mtd:mining}
	
\begin{figure}[]
		\centerline{\includegraphics[width=.9\columnwidth,trim=0 250 400 0, clip]{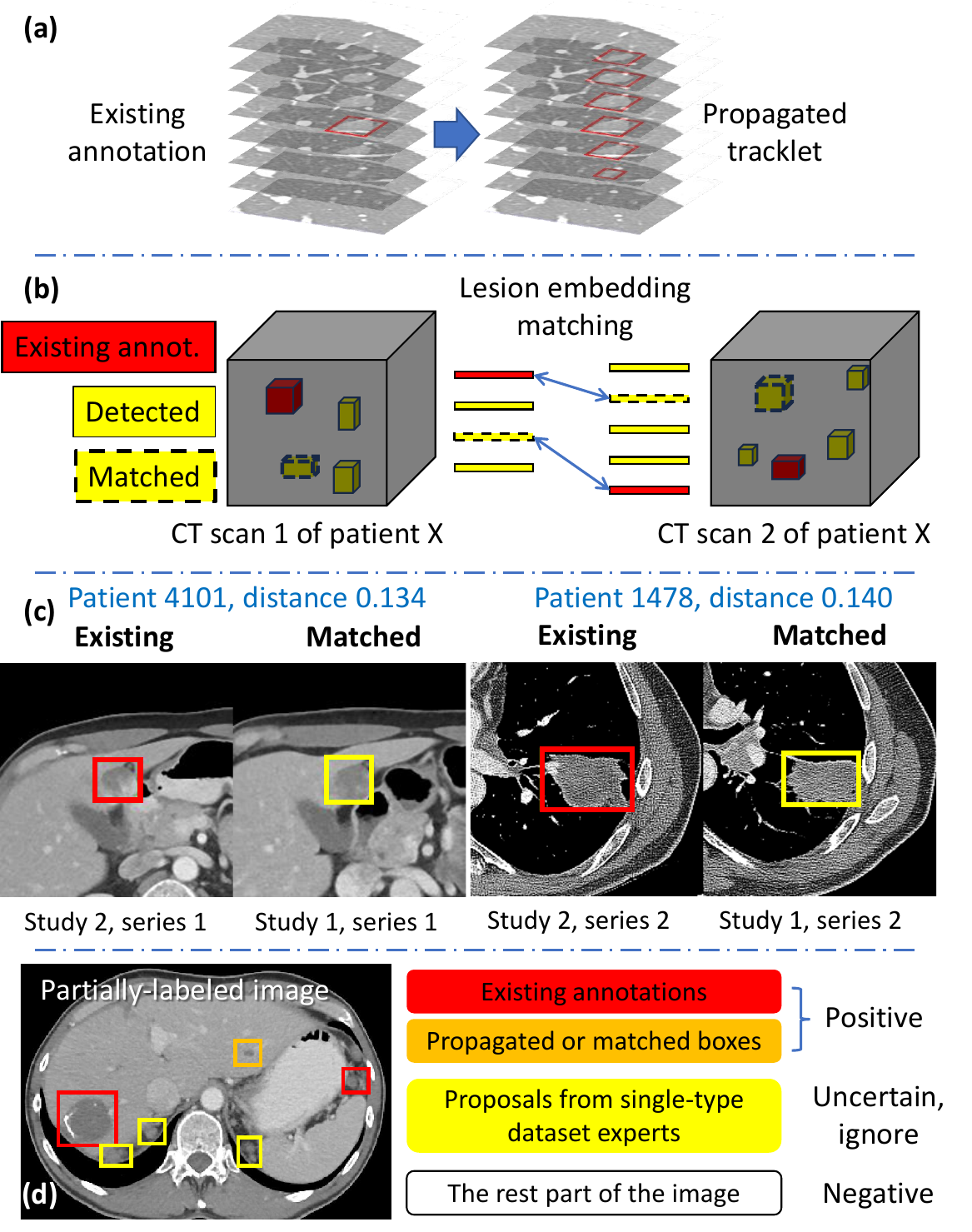}} 
		\caption{\textbf{(a)} Cross-slice box propagation from 2D lesion annotations. \textbf{(b)} Intra-patient lesion matching. \textbf{(c)} Examples of matched lesions within the same patient. Their embedding distance is also shown. \textbf{(d)} Cross-dataset lesion mining and the overall mining result.}
		\label{fig:miss_annot_mine}
	\end{figure}

	
	In this section, we introduce three strategies to mine missing annotations (MAs) and uncertain lesions in DeepLesion, as illustrated in \Fig{miss_annot_mine}. The ideas can be generalizable to other partially-labeled lesion datasets.
	
	\subsection{Cross-Slice Box Propagation}
	\label{subsec:mam:propa}
	In oncological practice, radiologists measure a 3D lesion on a 2D slice where it has the largest cross-sectional size according to the response evaluation criteria in solid tumours (RECIST)~\cite{Eisenhauer2009RECIST}, so the DeepLesion dataset only contains 2D boxes. We can find MAs by recovering the lesion's boxes in other slices~\cite{Wang2019missing}, see \Fig{miss_annot_mine} (a). We collect all 2D proposals on the training set of DeepLesion, and then group boxes in adjacent slices if their IoU is larger than $ \theta $. Given a set of 2D boxes $ A_i $ in slice $ i $, we check the boxes $ A_{i+1} $ in slice $ i+1 $. Denote $ a_{i,j} $ as the $ j $th box in slice $ i $. If $ a_{i,j}\in A_i,\, a_{i+1,k}\in A_{i+1} $, and $ \text{IoU}(a_{i,j}, a_{i+1,k}) > \theta$, then $ a_{i,j}$ and $a_{i+1,k} $ are considered as the same lesion and grouped together. It is also possible that one box cannot be grouped with any other boxes in adjacent slices. It will be treated as one group with a single 2D box. Thus, we can combine 2D boxes in a volume to several groups, also known as tracklets~\cite{Jin2018mining}, each representing a recovered 3D lesion proposal. Next, we check the ``key-slice IoU'' of a tracklet, namely, if any 2D box in the tracklet overlaps with an existing annotation with IoU $ >\theta $.  If so, the boxes on other slices in this tracklet are considered as mined MAs.
	
	\subsection{Intra-Patient Lesion Matching}
	
	Cross-slice box propagation leverages the spatial continuity of lesions, while intra-patient lesion matching utilizes their temporal continuity. In clinical practice, each patient generally undergo multiple CT scans (studies) at different time points to monitor their disease progress~\cite{Eisenhauer2009RECIST, Yan2018DeepLesion}. We find that within each patient, the same lesion may be annotated in one study but not another~\cite{Yan2018graph}. Therefore, we can establish correspondence between detected boxes and existing annotations to recover the unannotated MAs, see \Fig{miss_annot_mine} (b). Besides, each study typically contains multiple image volumes (series) that are scanned at the same time point but differ in reconstruction filters, contrast phases, etc. We can also mine MAs from different series similarly. We utilize the lesion embedding learned in LesaNet~\cite{Yan2019Lesa}, which encodes the body part, type, and attributes of lesions and has proved its efficacy in lesion retrieval. Specifically, in DeepLesion~\cite{Yan2018DeepLesion}, the patient ID, study ID, and series ID of each image volume are provided. We first collect all annotations and tracklets of one patient, then extract their lesion embeddings using LesaNet. LesaNet operates on 2D boxes, so we run it on the center slice of each tracklet. The distance of two embeddings should be small if they are from the same lesion instance. Hence, within each patient, we compute the L2 distance in the embedding space between every annotation and every tracklet (in a different volume) and keep those pairs whose distance is smaller than a threshold $ \delta $. \Fig{miss_annot_mine} (c) illustrates two pairs of matched lesions. Note that the matched MAs have similar but not identical appearance with existing annotations, since they are different in time point, contrast phase, etc. Therefore, the matched MAs can still provide useful new information when they are added in training.
	
	The intra-patient lesion matching strategy relies on the availability of multiple scans from the same patient. In real-world data collected from hospitals, follow-up scans and multiple image series generally exist, especially from cancer patients. Therefore, our strategy is expected to be useful. It can also save doctors' annotation time---doctors can annotate a lesion on one scan, and then multiple annotations from other series and studies can be mined.
	
	\subsection{Cross-Dataset Lesion Mining}
	\label{subsec:mam:CrossMine}

	The two strategies above cannot find an MA if it does not match with any existing annotation. Our solution is to explore the semantic overlap between datasets and distill knowledge from the single-type datasets. Recall that LENS is an ensemble of multiple dataset experts and can output several groups of proposals. Our intuition is that the single-type proposals generally have higher recall and fewer FPs in their specialties compared to the DeepLesion expert. This has been discussed in Sec.~\ref{subsec:mtd:LENS} and \Fig{DL_spec_det_cmp}. Therefore, for each 2D proposal from the single-type experts, if its detection score is higher than a threshold $ \sigma $ and it does not overlap with existing or mined annotations, we regard the proposal as a suspicious or uncertain lesion. It can either be an unlabeled lesion or a false positive (FP), so we choose to ignore them (exclude them in the loss function) during the retraining of LENS, see \Fig{miss_annot_mine} (d). Compared to treating them as true lesions, ignoring them prevents the FPs from injecting noise to the retraining process. The lesions mined by cross-slice box propagation and intra-patient lesion matching have high confidence to be true lesions, because strong prior knowledge (spatial and temporal continuity of lesions) is leveraged. Thus, we can treat them as true lesions during retraining. 
	
	Previous ULD algorithms~\cite{Wang2019universal, Li2019MVP, Yan2019MULAN} were all limited to the 22K labeled training slices in DeepLesion. It will bias the algorithms toward lesion-rich body parts and cause many FPs in under-represented body parts. With the three mining strategies in this section, we can mine MAs and uncertain lesions from the massive unlabeled slices to incorporate them in training and improve performance on the whole body.
	
	\section{Experiments}
	\label{sec:exp}
	
	\subsection{Data}
	\label{subsec:data}
	
	\begin{table}[]
		\begin{center}
			\scriptsize
			\setlength{\tabcolsep}{4pt}
			\renewcommand{\arraystretch}{1.3}
			\caption{Statistics of the four lesion datasets used in our work}
			\label{tbl:datasets}
			\begin{tabular}{p{1.5cm}p{.9cm}p{.8cm}p{1.2cm}p{.7cm}p{.8cm}p{1cm}}
				\hline
				Name	& Lesion types	& Organs	& \# 3D Volumes	& \# 2D Slices	& \# Lesions	& Fully-annotated?	 \\
				\hline
				DeepLesion~\cite{Yan2018DeepLesion}	& Various	& Whole body	& 10,594 sub-volumes	& 928K	& 32,735	& No \\
				LUNA~\cite{Setio2017LUNA}	& Lung nodule	& Lung	& 888	& 226K	& 1,186	& Yes \\
				LiTS~\cite{Bilic2019LiTS}	& Liver tumor	& Liver	& 131	& 85K	& 908	& Yes \\
				NIH-LN~\cite{NIH_LN_dataset}	& Lymph node	& Lymph node	& 176	& 134K	& 983	& Yes \\			
				\hline
			\end{tabular}
		\end{center}
	\end{table}
	
	\Table{datasets} lists the datasets used in this paper. DeepLesion~\cite{Yan2018DeepLesion} is a large universal lesion dataset containing 32,735 lesions annotated on 10,594 studies of 4,427 patients. It was mined from the National Institutes of Health Clinical Center based on radiologists' routine marks to measure significant image findings~\cite{Eisenhauer2009RECIST}. Thus, it closely reflects clinical needs. The LUNA (LUng Nodule Analysis) dataset~\cite{Setio2017LUNA} consists of 1,186 lung nodules annotated in 888 CT scans. LiTS (LIver Tumor Segmentation Benchmark)~\cite{Bilic2019LiTS} includes 201 CT scans with 0 to 75 liver tumors annotated per scan. We used 131 scans of them with released annotations. NIH-Lymph Node (NIH-LN)~\cite{NIH_LN_dataset} contains 388 mediastinal LNs on 90 CT scans and 595 abdominal LNs on 86 scans. Without loss of generality, we chose these three single-type datasets for joint learning with DeepLesion in this paper. 
	
	For DeepLesion, we used the official training set for training. The official test set includes only 2D slices and may contain missing annotations, which will bias the accuracy. We invited a board-certified radiologist to further comprehensively annotate 1000 sub-volumes in the test set of DeepLesion using 3D bounding boxes. 200 of them were used for validation and 800 for testing. In this fully-annotated test set, there are 4,155 lesions in 34,114 slices. For LUNA, LiTS, and NIH-LN, we randomly used 80\% of each dataset for training and 20\% for validation. We tried to build a unified lesion detection framework and adopted the same image preprocessing and data augmentation steps~\cite{Yan2019MULAN} for all datasets. First, we normalized the image orientations of all datasets. Then, we rescaled the 12-bit CT intensity range to floating-point numbers in [0,255] using a single windowing (-1024--3071 HU) that covers the intensity ranges of the lung, soft tissue, and bone. Every axial slice was resized so that each pixel corresponds to 0.8mm. We interpolated in the $ z $-axis to make the slice intervals of all volumes to be 2mm. The black borders in images were clipped for computation efficiency. When training, we did data augmentation by randomly resizing each slice with a ratio of 0.8$ \sim $1.2 and randomly shifting the image and annotation by -8$ \sim $8 pixels in $ x $ and $ y $ axes.
	
	\begin{table*}[t]
		\begin{center}
			\caption{Results with different components of the proposed framework}
			\label{tbl:ablation}
			\begin{tabular}{rccccccccccccc}
				\noalign{\smallskip}\hline\noalign{\smallskip}
				Method	& AFP	& Multi-dataset	& Proposal fusion	& MAM	& FPR & FP@0.125	& 0.25	& 0.5	& 1	& 2	& 4	& 8 & Average \\ 
				\hline\noalign{\smallskip}
Baseline~\cite{Yan2019MULAN}	&            &            &            &            &          & 11.2	& 16.3	& 24.3	& 32.8	& 41.6	& 50.9	& 60.1	& 33.9 \\
(a)                             & \checkmark &            &            &            &          & 15.8	& 21.4	& 27.9	& 35.9	& 43.4	& 52.0	& 60.9	& 36.8 \\
(b)                             & \checkmark & \checkmark &            &            &          & 14.3	& 21.5	& 28.2	& 35.1	& 44.4	& 53.9	& 63.4	& 37.3 \\
(c)                             & \checkmark & \checkmark & \checkmark &            &          & 15.9	& 22.8	& 30.1	& 37.7	& 46.7	& 56.6	& 66.1	& 39.4 \\
(d)                             &  & \checkmark & \checkmark & \checkmark &          & 18.3	& 26.3	& 34.1	& 44.8	& 55.5	& 65.4	& 75.4	& 45.7 \\
(e)                             & \checkmark &  &  & \checkmark &          & 22.0	& 28.4	& 36.6	& 45.2	& 55.0	& 65.5	& 75.0	& 46.8 \\
(f)                             & \checkmark & \checkmark &  & \checkmark &          & 21.3	& 28.3	& 37.1	& 46.7	& 55.5	& 66.2	& 75.9	& 47.3 \\
(g)                             & \checkmark & \checkmark & \checkmark & \checkmark &          & 21.6	& 29.9	& 37.6	& 46.7	& 56.7	& 65.8	& 75.3	& 47.6 \\
(h)                             & \checkmark & \checkmark & \checkmark & \checkmark & \checkmark & \bf 23.7	& \bf 31.6	& \bf 40.3	& \bf 50.0	& \bf 59.6	& \bf 69.5	& \bf 78.0	& \bf 50.4 \\
				
				\hline\noalign{\smallskip}
				\multicolumn{14}{p{.95\textwidth}}{Sensitivity (\%) at different FPs per sub-volume on the manually labeled test set of DeepLesion is shown. AFP: Anchor-free proposal network; MAM: Three missing annotation mining strategies in Sec.~\ref{sec:mtd:mining}; FPR: 3D false positive reduction network.}
			\end{tabular}
		\end{center}
	\end{table*}
	
	\subsection{Implementation}
	\label{subsec:implementation}
	
	LENS was implemented in PyTorch based on the maskrcnn-benchmark project~\cite{massa2018mrcnn}. The backbone of LENS were initialized with an ImageNet pretrained model. We used rectified Adam (RAdam)~\cite{liu2019radam} to train LENS for 8 epochs. We set the base learning rate to 0.0001, then reduced it to 1/10 after the 4th and 6th epochs. It took LENS 54ms to process a slice during inference on a Quadro RTX 6000 GPU. When training LENS, each batch had 4 images sampled from the same dataset~\cite{Wang2019universal}, where each image consisted of 9 axial CT slices for 3D feature fusion~\cite{Yan2019MULAN}. The training data in each dataset contained positive slices (with existing annotations or mined MAs) and randomly sampled negative slices (without lesions). Their ratio is 2:1 in each epoch. Since the datasets have different sizes, we tried to reweight the samples in the loss function, but no improvement was observed. For MA mining, we empirically chose the distance threshold for intra-patient lesion matching as $ \delta=0.15 $, the detection score threshold for cross-dataset lesion mining as $ \sigma=0.5 $.
	
	The 3D FPR was initialized with ImageNet pretrained ResNet-18 using inflated 3D~\cite{Carreira2017I3D}.  We used RAdam to train it for 6 epochs. We set the base learning rate to 0.0002, then reduced it to 1/10 after the 2nd and 4th epochs. The batch size was 32. When training FPR, the positive and negative samples were the TP and FP proposals of LENS on the training set of DeepLesion. If the key-slice IoU (Sec.~\ref{subsec:mam:propa}) of a proposal and an annotation or MA is larger than $ \theta $, it is considered a TP. It is an FP if its IoU is smaller than $ \theta_{\text{FP}} $ with any annotation, MA, or uncertain lesion. We used $ \theta=0.5, \theta_{\text{FP}}=0.3 $. Note that all IoU threshold $ \theta $ in this paper was set to 0.5. The ratio of TP and FP is 1:2 in each epoch.
	
	\subsection{Metric}
	\label{subsec:metrics}
	
	The free-response receiver operating characteristic (FROC) curve is the standard metric in lesion detection~\cite{Setio2017LUNA, Yan2018DeepLesion, Shin2016TMICNN}. 
	Following the LUNA challenge~\cite{Setio2017LUNA}, sensitivities at $1/8, 1/4, 1/2, 1, 2, 4, 8$ FPs per sub-volume are computed to show the recall at different precision levels. The average of these are referred as average sensitivity. We noticed that sometimes the detector identified smaller parts of a large or irregular lesion with a big ground-truth box (see \Fig{qualitative} (b) column 1 for an example). Although the IoU may be not high in such cases, the detection may still be viewed as a TP as it can also help the radiologists~\cite{Yan20183DCE}. To this end, we utilized the intersection over the detected bounding-box area ratio (IoBB) instead of IoU in evaluation. If the 3D IoBB of a proposal and a ground-truth is larger than 0.3, it is considered as a TP.
	
	\subsection{Results on DeepLesion}
	
	\Table{ablation} displays our \textbf{main results} on the 800 manually labeled sub-volumes in the test set of DeepLesion. The baseline is the previous state-of-the-art method on DeepLesion, MULAN~\cite{Yan2019MULAN}, which was trained on the official annotations of DeepLesion alone and tested on the new test set. First, we replaced the region proposal network (RPN)~\cite{Ren2015faster} in MULAN with our proposed anchor-free proposal network (AFP) (row {\bf(a)}). The accuracy improved, which is mainly because AFP used focal loss to learn from the whole image to distinguish between lesions and non-lesions, whereas RPN sampled limited positive and negative anchors. In row {\bf(b)}, we added the three single-type datasets for multi-task joint training, the average sensitivity was further improved by 0.5\%. The shared backbone and FC layers in LENS can learn better feature representation through multi-source CT datasets. The improvement is not very prominent possibly because the DeepLesion dataset is already large. Experiments in the next section show that joint training improves accuracy on small datasets significantly. The results in row (b) only used the output of the DeepLesion dataset expert. In row {\bf(c)}, we further fused the proposals of the four dataset experts. The predictions of different dataset experts are complementary (\Fig{DL_spec_det_cmp}). Therefore, fusing them improved ULD sensitivity. Rows (d)--(g) all employed the MA mining strategies in Sec.~\ref{sec:mtd:mining}, which significantly improved the accuracy. In row {\bf(d)}, AFP was replaced with FCOS~\cite{Tian2019FCOS}. In row {\bf(e)}, only the DeepLesion dataset was used in training. In row {\bf(f)}, the three single-type datasets were used in training but their proposals were not fused with DeepLesion. In row {\bf(g)}, all components of LENS were applied. Finally, in row {\bf(h)}, cascading a 3D FPR classifier obtained another accuracy gain of 2.8\%. Using the FPR score alone without averaging it with the detection score (\Eq{final_lesion_score}) is not promising (average sensitivity 32.0\%). It is possibly because our detection framework is more powerful in differentiating lesions from non-lesions, so its score should not be ignored. FPR scores are complementary to detection scores since FPR is trained using hard negative proposals of LENS, so averaging the two scores improved the performance. Our framework is scalable and can easily incorporate more datasets with heterogeneous labels, which we will investigate in the future.
	
\begin{table}[h]
	\begin{center}
		\caption{Comparison of different proposal networks}
		\label{tbl:prop_net}
		\begin{tabular}{lccc}
			\noalign{\smallskip}\hline\noalign{\smallskip}
			Method	& \# Stage & Average sensitivity	& Inference time (ms) \\
			\hline\noalign{\smallskip}
			RPN~\cite{Ren2015faster}	& two & 42.5	& 62 \\
			CenterNet~\cite{Zhou2019CenterNet}  & two & 45.3  & 59 \\
			FCOS~\cite{Tian2019FCOS}	& one & 41.7	& \bf 29 \\
			FCOS	& two & 45.7	& 54 \\
			AFP	& two & \bf 47.6	& 54 \\
			\hline
		\end{tabular}
	\end{center}
\end{table}
	
	\Table{prop_net} compares the accuracy and inference time per slice of different \textbf{proposal networks}. The AFP in LENS (row (d) in \Table{ablation}) was replaced with other parts unchanged. AFP obtained better accuracy than the other three algorithms. AFP divides a box into positive, ignore, and negative regions during training (Sec.~\ref{subsec:mtd:AFP}), which brings more positive supervision information than CenterNet~\cite{Zhou2019CenterNet}. It is beneficial since lesions are sparse in the image. Compared to FCOS~\cite{Tian2019FCOS}, AFP learns from the informative central part of a box~\cite{Tian2019FCOS} without the need for a center-ness branch. 
	We also tried to use FCOS as the proposal network and remove the detection head of LENS to make it a one-stage detector. It was faster but less accurate than its two-stage counterpart, indicating the importance of the detection head.
	
	\begin{table}[h]
		\begin{center}
			\setlength{\tabcolsep}{3pt}
			\caption{Comparison of the missing annotation mining strategies}
			\label{tbl:MAM}
			\begin{tabular}{lp{1.2cm}p{1.1cm}p{1.1cm}p{1.2cm}p{1.2cm}}
				\noalign{\smallskip}\hline\noalign{\smallskip}
				Method	& Average sensitivity	& \# 3D GT	& \# 2D GT	& \# 2D uncertain	& \# Training slices \\
				\hline\noalign{\smallskip}
				No mining	& 36.8	& 22.8K	& 22.8K	& 0	& 22.4K \\
				+ Cross-slice	& 37.7	& 22.8K	& 186K	& 0	& 163K \\
				+ Intra-patient	& 40.7	& 34.3K	& 250K	& 0	& 192K \\
				+ Cross-dataset	& 44.6	& 34.3K	& 250K	& 117K	& 192K \\
				+ Unlabeled		& \bf 46.8	& 34.3K	& 250K	& 442K	& 646K \\
				\hline\noalign{\smallskip}
				Wang \etal~\cite{Wang2019missing}	& 38.0	& --	& 56.8K	& 0	& 53.4K \\
				Data distillation~\cite{Radosavovic2017data}	& 43.7	& --	& 511K	& 0	& 646K \\
				\hline\noalign{\smallskip}
			\end{tabular}
		\end{center}
	\end{table}
	
	\begin{figure}[h]
		\centerline{\includegraphics[width=\columnwidth,trim=0 0 0 0, clip]{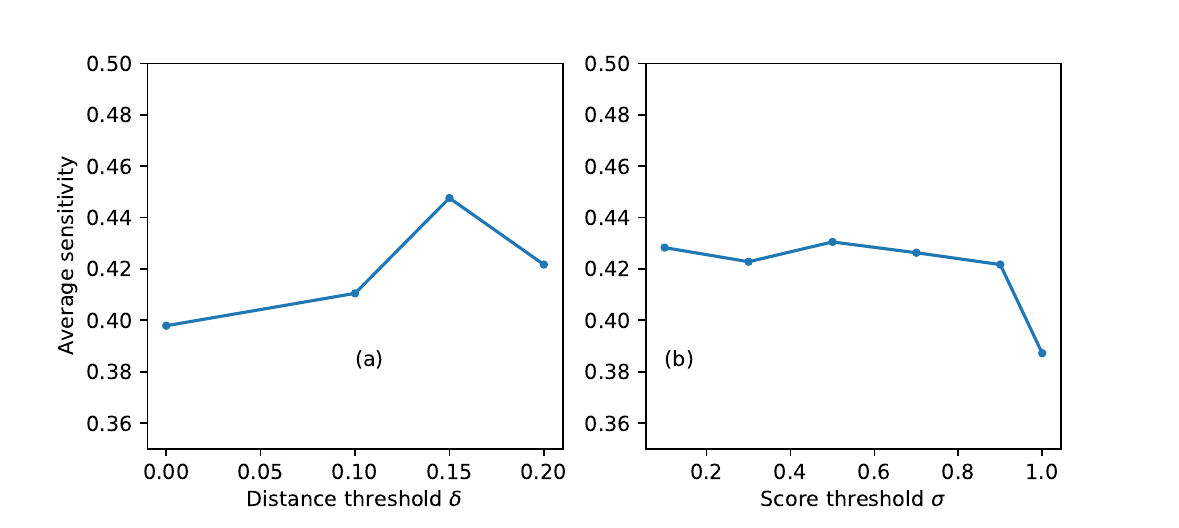}} 
		\caption{Average sensitivity on the validation set of DeepLesion with different hyper-parameters. The distance threshold $ \delta $ for intra-patient lesion matching and the detection score threshold $ \sigma $ for cross-dataset lesion mining are tuned. If $ \delta=0 $, no matched lesion is added in retraining. If $ \sigma=1 $, no uncertain lesion is ignored during retraining.}
		\label{fig:hyper_param}
	\end{figure}
	
	In \Table{MAM}, we evaluate the three \textbf{MA mining strategies}: cross-slice box propagation, intra-patient lesion matching, and cross-dataset lesion mining. To eliminate the influence of multiple datasets, we trained LENS on DeepLesion alone. When no mining was applied, the model learned from the original 22.8K 2D lesion ground-truths (GTs) on 22.4K slices in the training set of DeepLesion. Cross-slice box propagation can infer more 2D boxes in the adjacent slices of existing lesion GTs, which brought an accuracy gain of 0.9\%. Note that we randomly sampled one positive slice for each 3D GT in training. 
	Intra-patient lesion matching discovered 11.5K new lesion instances (\# 3D GT), which further improved the accuracy by 3\%. We randomly checked 100 of the new instances and found 90\% are true lesions. During cross-dataset lesion mining, 117K uncertain 2D boxes were mined by the single-type dataset experts. Examples of the uncertain lesions can be found in \Fig{DL_spec_det_cmp}, \ie~the boxes that were detected by single-type dataset experts but missed by DeepLesion. Accuracy was increased by 3.9\% by ignoring these boxes during training. The ``unlabeled'' in the last row of \Table{MAM} indicates the image slices containing neither ground-truth lesions nor mined missing annotations. To leverage these unlabeled slices in training, we apply cross-dataset lesion mining on them to find uncertain lesions. Sampling these images in training while ignoring the uncertain area brought a 2.2\% accuracy gain since they can make the model learn more about normal appearances in the whole body. We also tried to use the retrained LENS to mine MAs for one more round and retrain LENS again. Although 1.4K more MAs were discovered, the improvement on average sensitivity is less than 0.2\%. It is possibly because there are little new information in these new MAs. Therefore, we decided not to run more iterations. The impact of the hyper-parameter values is studied in \Fig{hyper_param}.
	
	For comparison, we also tested two recent algorithms that can handle MAs in detection, see \Table{MAM}. Wang \etal~\cite{Wang2019missing} found MAs by checking the adjacent slices of the original annotations. The result was mildly improved. Data distillation was proposed in~\cite{Radosavovic2017data}, where predictions from multiple transformations of unlabeled data were merged to generate new training annotations. We followed~\cite{Radosavovic2017data} and did data augmentation in both training and testing, including random flipping in three axes, random scaling in the range of (0.5, 1.5), and random translation between (-8, 8) pixels. In testing, we augmented each image 3 times and fused the predicted boxes by weighted averaging, which were further treated as mined MAs and used for retraining. This strategy was effective and improved the accuracy to 43.7\%, but the mined MAs may contain many false positives. Our methods leveraged multiple prior knowledge, so the mined MAs are more reliable and lead to better accuracy.
	
\begin{table}[h]
	\begin{center}
		\setlength{\tabcolsep}{3pt}
		\caption{Different strategies to combine multiple datasets}
		\label{tbl:multi_dataset}
		\begin{tabular}{lccc}
			\noalign{\smallskip}\hline\noalign{\smallskip}
			Method	& Avg. sensitivity   & Infer. time    & Model size \\
			\hline\noalign{\smallskip}
			Single dataset	& 36.8 	& 1$\times$	& 1$\times$ \\ 
			Data pooling~\cite{Lenga2020Continual}	& 39.4 	& 1$\times$	& 1$\times$ \\
			Positive data pooling	& 32.8 	& 1$\times$	& 1$\times$ \\ 
			Separate models & 39.6	& 4$\times$	& 4$\times$ \\
			Proposed	& \bf 47.6 	& 1.8$\times$	& 1.1$\times$ \\ 
			Ignored as GT	& 44.9 	& 1.8$\times$	& 1.1$\times$ \\
			Proposed + domain adapt. \cite{Wang2019universal}   & 47.1  & 1.9$\times$  & 1.1$\times$  \\ 
			\hline
		\end{tabular}
	\end{center}
\end{table}

	Several strategies to \textbf{combine multiple lesion datasets} are compared in \Table{multi_dataset}. They all used the same backbone and AFP. ``Single dataset'' learned from DeepLesion alone. ``Data pooling'' directly pooled DeepLesion and single-type datasets and treated them as one task. ``Positive data pooling'' only sampled positive regions from the single-type datasets to joint train with DeepLesion to avoid the influence of MAs of other types. We find data pooling improved upon single dataset but positive data pooling is actually worse, which may be because the positive samples from single-type datasets contain concept shift~\cite{Cohen2020domain} relative to DeepLesion. Data pooling further added lots of negative samples from single-type datasets. Although there may be some MAs of other types, they may still be helpful for the model to learn the appearance of normal tissues from multi-source CT images, so as to reduce FPs and improve the sensitivity at low FP levels. This is also why leveraging the unlabeled slices in DeepLesion is useful. ``Separate models'' learned a detector for each dataset and fused their proposals in inference. It is the slowest approach with the largest model size. Its accuracy is better than single dataset but worse than our proposed framework, possibly because each separate model performed no better than the corresponding dataset expert in our joint learning model. Our proposed framework performed the best by integrating knowledge of multiple datasets through feature sharing, proposal fusion, and annotation mining. We also find that treating the mined uncertain lesions as ignored is better than regarding them as true lesions, possibly because they contain some noise and concept shift. The domain adaptation module~\cite{Wang2019universal} assigns dataset-specific feature attentions, but it did not improve probably because all datasets we used are CT images with small difference in the image domain.
	
	\begin{figure*}[t]
		\centerline{\includegraphics[width=\textwidth,trim=0 370 0 0, clip]{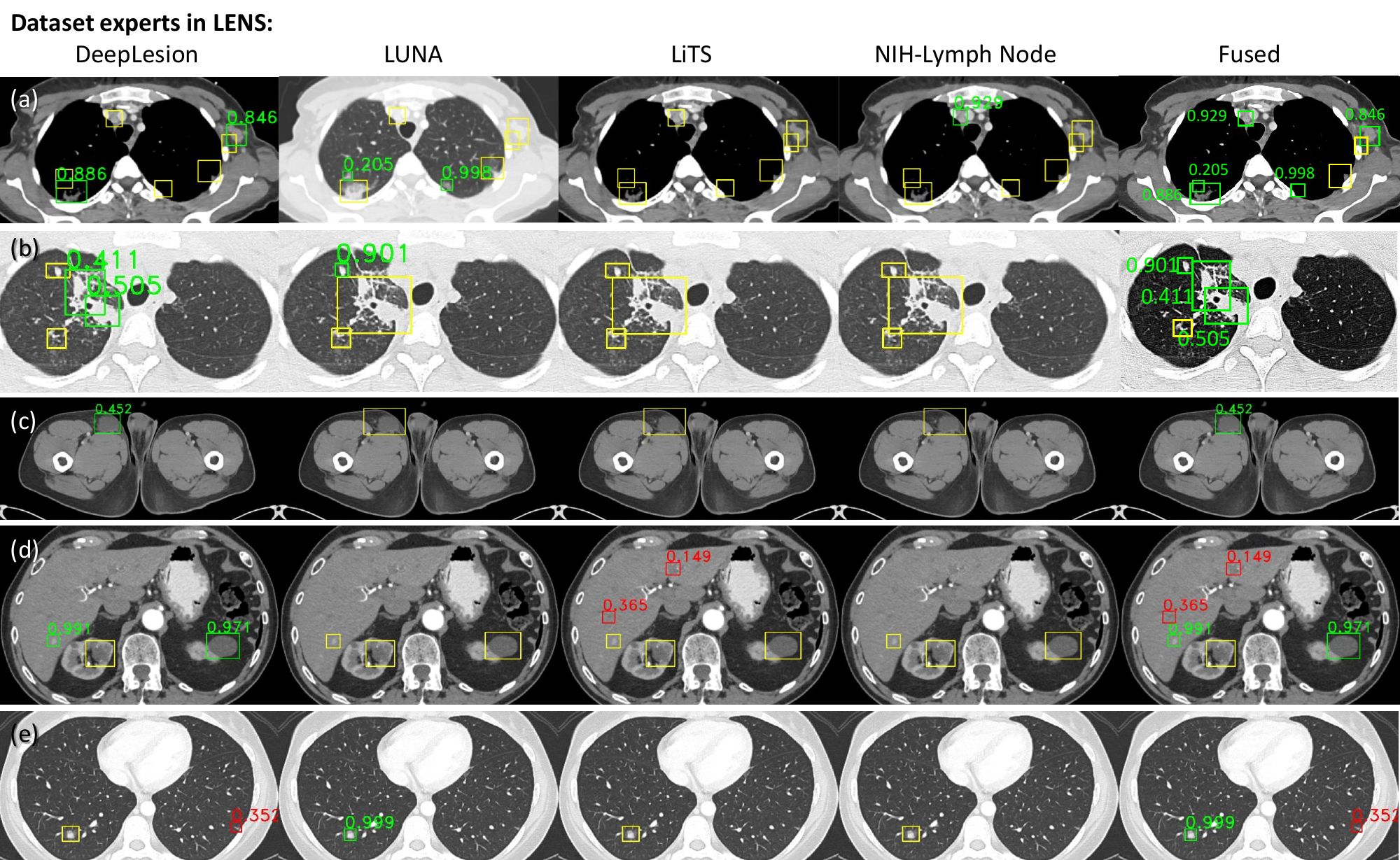}} 
		\caption{Exemplar detection results of our proposed framework on the test set of DeepLesion. Columns 1--4 are the proposals from the four dataset experts of LENS. Dataset expert $ i $ corresponds to the $ i $th output of the multi-task detection head in \Fig{LENS_framework} that is trained on the $ i $th dataset. Column 5 shows the fused proposals of the four dataset experts. Green, red, and yellow boxes indicate true positives (TP), false positives (FP), and false negatives (FN), respectively. Note that an FN box is a manual 3D ground-truth annotation in a 2D slice, thus may be slightly larger than the actual lesion in this slice. The detection scores are also displayed. We show boxes with scores greater than 0.1. Different intensity windows (soft tissue, lung) are used depending on the lesions to show. It can be found that the dataset experts are complementary. Best viewed in color.}
		\label{fig:qualitative}
	\end{figure*}
	
	\textbf{Qualitative results} are displayed in \Fig{qualitative}. 
	It is clear that the predictions of different dataset experts are complementary and fusing them can improve the final sensitivity. The single-type experts are able to detect difficult cases that are in their specialty, such as small lung nodules and lymph nodes (subplots (a)(b)(e)) and indistinct liver tumors (\Fig{DL_spec_det_cmp} subplot (d)) that may be missed by the universal expert. But lesions of certain appearances, sizes, or contrast phases can be uncommon in the single-type datasets, thus will be missed by the single-type experts even if they are in their specialties. The universal dataset, on the other hand, contains more diverse training samples, thus can detect more lesion types (\eg~the inguinal lymphocyst in (c), and kidney lesion in (d)), as well as those missed by the single-type experts (\eg~the large lung lesions in (a)(b), axillary LN in (a), and liver lesion in (d)). 
	Subplots (d)(e) illustrate possible issues in both single-dataset and multi-dataset learning. The single-type datasets may introduce some FPs (\eg~the LiTS expert in subplot (d)), which is possibly because the distribution discrepancy across datasets (patient population, contrast phase, etc.) makes similar image appearance has different meanings in LiTS and DeepLesion. How to deal with this discrepancy remains a question. In subplot (e), the DeepLesion expert detected an FP, while the LUNA expert made the right decision. Can we make the model realize that the LUNA expert is more reliable in this case? Simply assigning high weights to LUNA on all lung lesions is problematic because there are many lung lesions that LUNA cannot detect (subplots (a)(b)). Future work may include training a gating head to differentiate different dataset experts' specialties. 
	
	\subsection{Results on Single-Type Datasets}
	
	\begin{figure}[]
		\centerline{\includegraphics[width=\columnwidth,trim=0 0 0 0, clip]{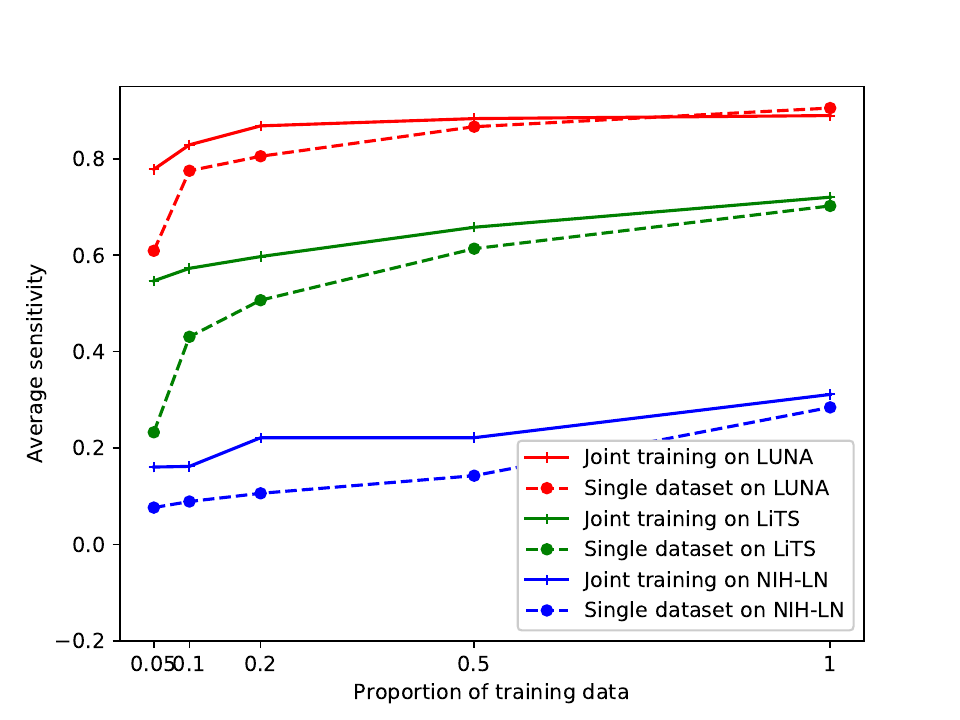}} 
		\caption{Comparison of single-dataset learning and multi-dataset joint training with different proportions of training data in the single-type datasets. We report the average sensitivity at $ \frac{1}{8}{\small\sim}8 $ FPs per volume~\cite{Setio2017LUNA}.}
		\label{fig:single_type_acc}
	\end{figure}
	
	The joint training strategy in LENS can improve not only DeepLesion, but also the single-type datasets. We combined DeepLesion with a proportion of training volumes from all single-type datasets to train LENS. For comparison, we trained LENS with one single-type dataset each time of the same training size. Evaluation was made on the validation set (20\% of each dataset). \Fig{single_type_acc} shows that joint training always outperformed single dataset, especially when the number of training samples is smaller. The only exception is LUNA with 100\% training data. This is because the diverse lesions in DeepLesion is helpful in learning effective CT image features and avoiding overfitting. It is particularly useful in medical image analysis where training data is often limited. It also indicates that the network has the capacity to learn different lesion types in multiple datasets at the same time. Among the three single-type datasets, lung nodules have relatively distinct appearance, thus are easier to learn. Besides, LUNA has the most training data, so the superiority of joint training is smaller. Some liver tumors have clear separation with normal tissues, while others can be subtle, making it a harder task. Lymph nodes exist throughout the body and are often hard to be discriminated from the surrounding vessels, muscles, and other organs, leading to the lowest accuracy~\cite{Zhu2020gating, Zhu2020scatter}.
	
	Note that we aim to compare single-dataset with joint learning but not to compare LENS with existing best algorithms specially designed on each single-type dataset. As a supplemental experiment to demonstrate the detection ability of LENS, we ran our model using the official LUNA evaluation metric~\cite{Setio2017LUNA} and compared it with other methods in the challenge. Without further tuned parameters and tricks such as organ masking and model ensemble, LENS obtained an average sensitivity of 0.898, ranking 6th in the leaderboard~\cite{LUNA16Res}, between the 5th team zhongliu\_xie (0.922) and the 6th team iDST-VC (0.897).
	\section{Conclusion}
	\label{sec:conclusion}
	
	In this paper, we studied two key problems in annotation-efficient deep learning: learning with multiple heterogeneous datasets and learning with partial labels, for the lesion detection task. We proposed lesion ensemble (LENS) to jointly learn from multiple datasets and integrate their knowledge through feature sharing and proposal fusion. Strategies are designed to mine missing annotations from partially-labeled datasets by exploiting clinical prior knowledge and cross-dataset knowledge transfer. 
	Our framework provides a powerful means to exploit multi-source, heterogeneously and imperfectly labeled data, significantly pushing forward the performance of universal lesion detection.
	
	In the proposed LENS detector, we pool the proposals from all dataset experts to enhance the sensitivity for universal lesion detection. However, multiple dataset experts may also generate false positives (FPs), as shown in Fig.~6 (d)(e). Currently, we use another false positive reduction classifier to reduce FPs. It may be better if we can include a gating head in LENS to differentiate the specialties of the dataset experts. For example, if the proposal is a small lung nodule, the gating head should know that the detection score from the LUNA head is more reliable. In this way, the predictions from multiple experts can be combined more intelligently.
	
	\bibliographystyle{IEEEtran}
	\bibliography{tmi}

\begin{thebibliography}{10}
\providecommand{\url}[1]{#1}
\csname url@samestyle\endcsname
\providecommand{\newblock}{\relax}
\providecommand{\bibinfo}[2]{#2}
\providecommand{\BIBentrySTDinterwordspacing}{\spaceskip=0pt\relax}
\providecommand{\BIBentryALTinterwordstretchfactor}{4}
\providecommand{\BIBentryALTinterwordspacing}{\spaceskip=\fontdimen2\font plus
\BIBentryALTinterwordstretchfactor\fontdimen3\font minus
  \fontdimen4\font\relax}
\providecommand{\BIBforeignlanguage}[2]{{%
\expandafter\ifx\csname l@#1\endcsname\relax
\typeout{** WARNING: IEEEtran.bst: No hyphenation pattern has been}%
\typeout{** loaded for the language `#1'. Using the pattern for}%
\typeout{** the default language instead.}%
\else
\language=\csname l@#1\endcsname
\fi
#2}}
\providecommand{\BIBdecl}{\relax}
\BIBdecl

\bibitem{Tajbakhsh2020imperfect}
N.~Tajbakhsh, L.~Jeyaseelan, Q.~Li, J.~Chiang, Z.~Wu, and X.~Ding, ``{Embracing
  Imperfect Datasets: A Review of Deep Learning Solutions for Medical Image
  Segmentation},'' \emph{Medical Image Analysis}, 2020.

\bibitem{Huang2019U2}
C.~Huang, H.~Han, Q.~Yao, S.~Zhu, and S.~K. Zhou, ``{3D U$^2$-Net: A 3D
  Universal U-Net for Multi-Domain Medical Image Segmentation},'' in
  \emph{Proc. Int. Conf. Med. Image Comput. Computer-Assisted Intervent.},
  2019.

\bibitem{Zhou2019organ}
Y.~Zhou, Z.~Li, S.~Bai, C.~Wang, X.~Chen, M.~Han, E.~Fishman, and A.~Yuille,
  ``{Prior-aware Neural Network for Partially-Supervised Multi-Organ
  Segmentation},'' in \emph{Proc. IEEE Int. Conf. Comput. Vis.}, 2019.

\bibitem{Luo2020imperfect}
L.~Luo, L.~Yu, H.~Chen, Q.~Liu, X.~Wang, J.~Xu, and P.-A. Heng, ``{Deep Mining
  External Imperfect Data for Chest X-ray Disease Screening},'' \emph{IEEE
  Trans. Med. Imaging}, 2020.

\bibitem{Cohen2020domain}
\BIBentryALTinterwordspacing
J.~P. Cohen, M.~Hashir, R.~Brooks, and H.~Bertrand, ``{On the limits of
  cross-domain generalization in automated X-ray prediction},'' in
  \emph{Medical Imaging with Deep Learning}, 2020, pp. 1--13. [Online].
  Available: \url{https://github.com/ieee8023/xray-generalization}
\BIBentrySTDinterwordspacing

\bibitem{Dmitriev2019seg}
K.~Dmitriev and A.~E. Kaufman, ``{Learning Multi-Class Segmentations From
  Single-Class Datasets},'' in \emph{Proc. IEEE Conf. Comput. Vis. Pattern
  Recognit.}, 2019, pp. 9501--9511.

\bibitem{Gundel2019Location}
S.~G{\"{u}}ndel, S.~Grbic, B.~Georgescu, S.~Liu, A.~Maier, and D.~Comaniciu,
  ``{Learning to recognize abnormalities in chest X-rays with location-aware
  dense networks},'' in \emph{Iberoam. Congr. Pattern Recognit.}, vol. 11401
  LNCS, 2019, pp. 757--765.

\bibitem{Wang2017ChestXray}
X.~Wang, Y.~Peng, L.~Lu, Z.~Lu, M.~Bagheri, and R.~M. Summers, ``{ChestX-ray8:
  Hospital-scale Chest X-ray Database and Benchmarks on Weakly-Supervised
  Classification and Localization of Common Thorax Diseases},'' in \emph{Proc.
  IEEE Conf. Comput. Vis. Pattern Recognit.}, 2017, pp. 2097--2106.

\bibitem{Yan2018DeepLesion}
K.~Yan, X.~Wang, L.~Lu, and R.~M. Summers, ``{DeepLesion: automated mining of
  large-scale lesion annotations and universal lesion detection with deep
  learning},'' \emph{Journal of Medical Imaging}, vol.~5, no.~3, 2018.

\bibitem{Yan2018graph}
K.~Yan~et al., ``{Deep Lesion Graphs in the Wild: Relationship Learning and
  Organization of Significant Radiology Image Findings in a Diverse Large-scale
  Lesion Database},'' in \emph{Proc. IEEE Conf. Comput. Vis. Pattern
  Recognit.}, 2018.

\bibitem{Eisenhauer2009RECIST}
E.~A. Eisenhauer~et al., ``{New response evaluation criteria in solid tumours:
  Revised RECIST guideline (version 1.1)},'' \emph{European Journal of Cancer},
  vol.~45, no.~2, pp. 228--247, 2009.

\bibitem{Litjens2017survey}
G.~Litjens~et al, ``{A survey on deep learning in medical image analysis},''
  \emph{Medical Image Analysis}, vol.~42, pp. 60--88, dec 2017.

\bibitem{Sahiner2018survey}
B.~Sahiner~et al, ``{Deep learning in medical imaging and radiation therapy},''
  \emph{Med. Phys.}, oct 2018.

\bibitem{Setio2017LUNA}
A.~A.~A. Setio, A.~Traverso, and T.~de~Bel~et al., ``{Validation, comparison,
  and combination of algorithms for automatic detection of pulmonary nodules in
  computed tomography images: The LUNA16 challenge},'' \emph{Med. Image Anal.},
  vol.~42, pp. 1--13, 2017.

\bibitem{Zhu2018DeepEM}
W.~Zhu, Y.~S. Vang, Y.~Huang, and X.~Xie, ``{DeepEM: Deep 3D ConvNets With EM
  For Weakly Supervised Pulmonary Nodule Detection},'' in \emph{Proc. Int.
  Conf. Med. Image Comput. Computer-Assisted Intervent.}\hskip 1em plus 0.5em
  minus 0.4em\relax Springer, Cham, sep 2018, pp. 812--820.

\bibitem{dou2017multilevel}
Q.~Dou, H.~Chen, L.~Yu, J.~Qin, and P.~A. Heng, ``{Multilevel Contextual 3-D
  CNNs for False Positive Reduction in Pulmonary Nodule Detection},''
  \emph{IEEE Trans. Biomed. Eng.}, vol.~64, no.~7, pp. 1558--1567, 2017.

\bibitem{Wang2019LiTS}
X.~Wang, S.~Han, Y.~Chen, D.~Gao, and N.~Vasconcelos, ``{Volumetric Attention
  for 3D Medical Image Segmentation and Detection},'' in \emph{Proc. Int. Conf.
  Med. Image Comput. Computer-Assisted Intervent.}, 2019, pp. 175--184.

\bibitem{Roth2016randView}
H.~R. Roth, L.~Lu, J.~Liu, J.~Yao, A.~Seff, K.~Cherry, L.~Kim, and R.~M.
  Summers, ``{Improving Computer-Aided Detection Using Convolutional Neural
  Networks and Random View Aggregation},'' \emph{IEEE Trans. Med. Imaging},
  vol.~35, no.~5, pp. 1170--1181, 2016.

\bibitem{Shin2016TMICNN}
H.~C. Shin, H.~R. Roth, M.~Gao, L.~Lu, Z.~Xu, I.~Nogues, J.~Yao, D.~Mollura,
  and R.~M. Summers, ``{Deep Convolutional Neural Networks for Computer-Aided
  Detection: CNN Architectures, Dataset Characteristics and Transfer
  Learning},'' \emph{IEEE Trans. Med. Imaging}, vol.~35, no.~5, pp. 1285--1298,
  2016.

\bibitem{zhu2020detecting}
Z.~Zhu~et al., ``Detecting scatteredly-distributed, small, andcritically
  important objects in 3d oncologyimaging via decision stratification,''
  \emph{arXiv preprint arXiv:2005.13705}, 2020.

\bibitem{RSNA2020template}
\BIBentryALTinterwordspacing
RSNA, ``{RadReport Template Library},'' 2020. [Online]. Available:
  \url{https://radreport.org/}
\BIBentrySTDinterwordspacing

\bibitem{Yan20183DCE}
K.~Yan, M.~Bagheri, and R.~M. Summers, ``{3D context enhanced region-based
  convolutional neural network for end-to-end lesion detection},'' in
  \emph{Proc. Int. Conf. Med. Image Comput. Computer-Assisted Intervent.}, vol.
  11070 LNCS, 2018, pp. 511--519.

\bibitem{Wang2019universal}
X.~Wang, Z.~Cai, D.~Gao, and N.~Vasconcelos, ``{Towards Universal Object
  Detection by Domain Attention},'' in \emph{Proc. IEEE Conf. Comput. Vis.
  Pattern Recognit.}, 2019.

\bibitem{Li2019MVP}
Z.~Li, S.~Zhang, J.~Zhang, K.~Huang, Y.~Wang, and Y.~Yu, ``{MVP-Net: Multi-view
  FPN with Position-aware Attention for Deep Universal Lesion Detection},'' in
  \emph{Proc. Int. Conf. Med. Image Comput. Computer-Assisted Intervent.},
  2019.

\bibitem{Yan2019MULAN}
K.~Yan, Y.~Tang, Y.~Peng, V.~Sandfort, M.~Bagheri, Z.~Lu, and R.~M. Summers,
  ``{MULAN : Multitask Universal Lesion Analysis Network for Joint Lesion
  Detection , Tagging , and Segmentation},'' in \emph{Proc. Int. Conf. Med.
  Image Comput. Computer-Assisted Intervent.}, 2019.

\bibitem{NIH_LN_dataset}
\BIBentryALTinterwordspacing
``{CT Lymph Nodes dataset - The Cancer Imaging Archive (TCIA) Public Access},''
  2016. [Online]. Available:
  \url{https://wiki.cancerimagingarchive.net/display/Public/CT+Lymph+Nodes}
\BIBentrySTDinterwordspacing

\bibitem{Bilic2019LiTS}
\BIBentryALTinterwordspacing
P.~Bilic~et al., ``{The Liver Tumor Segmentation Benchmark (LiTS)},'' Tech.
  Rep., 2019. [Online]. Available: \url{http://arxiv.org/abs/1901.04056}
\BIBentrySTDinterwordspacing

\bibitem{Ren2015faster}
S.~Ren, K.~He, R.~Girshick, and J.~Sun, ``{Faster r-cnn: Towards real-time
  object detection with region proposal networks},'' in \emph{Proc. Advances
  Neural Inf. Process. Syst.}, 2015, pp. 91--99.

\bibitem{He2017MaskRCNN}
K.~He, G.~Gkioxari, P.~Dollar, and R.~Girshick, ``{Mask R-CNN},'' in
  \emph{Proc. IEEE Int. Conf. Comput. Vis.}, 2017, pp. 2980--2988.

\bibitem{Tang2019Uldor}
Y.~Tang, K.~Yan, Y.-X. Tang, J.~Liu, J.~Xiao, and R.~M. Summers, ``{ULDor: A
  Universal Lesion Detector for CT Scans with Pseudo Masks and Hard Negative
  Example Mining},'' in \emph{Proc. IEEE Int. Symposium Biomedical Imag.},
  2019.

\bibitem{Lenga2020Continual}
M.~Lenga, H.~Schulz, and A.~Saalbach, ``{Continual Learning for Domain
  Adaptation in Chest X-ray Classification},'' in \emph{Proc. Med. Image Deep
  Learn.}, 2020.

\bibitem{Hinton2014distill}
G.~Hinton, O.~Vinyals, and J.~Dean, ``{Distilling the Knowledge in a Neural
  Network},'' in \emph{Proc. Advances Neural Inf. Process. Syst.}, 2014, pp.
  1--9.

\bibitem{Radosavovic2017data}
I.~Radosavovic, P.~Doll{\'{a}}r, R.~Girshick, G.~Gkioxari, and K.~He, ``{Data
  Distillation: Towards Omni-Supervised Learning},'' in \emph{Proc. IEEE Conf.
  Comput. Vis. Pattern Recognit.}, 2017.

\bibitem{Niitani2019sample}
Y.~Niitani, T.~Akiba, T.~Kerola, T.~Ogawa, S.~Sano, and S.~Suzuki, ``{Sampling
  Techniques for Large-Scale Object Detection from Sparsely Annotated
  Objects},'' in \emph{Proc. IEEE Conf. Comput. Vis. Pattern Recognit.}, 2019.

\bibitem{Jin2018mining}
S.~Y. Jin~et al., ``{Unsupervised hard example mining from videos for improved
  object detection},'' in \emph{Proc. Eur. Conf. Comput. Vis.}, vol. 11217
  LNCS, aug 2018, pp. 316--333.

\bibitem{Cai2020harvest}
\BIBentryALTinterwordspacing
J.~Cai, A.~P. Harrison, Y.~Zheng, K.~Yan, Y.~Huo, J.~Xiao, L.~Yang, and L.~Lu,
  ``{Lesion Harvester: Iteratively Mining Unlabeled Lesions and Hard-Negative
  Examples at Scale},'' jan 2020. [Online]. Available:
  \url{http://arxiv.org/abs/2001.07776}
\BIBentrySTDinterwordspacing

\bibitem{Wang2019missing}
Z.~Wang, Z.~Li, S.~Zhang, J.~Zhang, and K.~Huang, ``{Semi-supervised lesion
  detection with reliable label propagation and missing label mining},'' in
  \emph{Chinese Conf. Pattern Recognit. Comput. Vis.}, vol. 11858 LNCS, 2019,
  pp. 291--302.

\bibitem{Pan2010transfer}
S.~J. Pan and Q.~Yang, ``{A survey on transfer learning},'' \emph{IEEE
  Transactions on Knowledge and Data Engineering}, vol.~22, no.~10, pp.
  1345--1359, 2010.

\bibitem{Zhou2020survey}
\BIBentryALTinterwordspacing
S.~K. Zhou, H.~Greenspan, C.~Davatzikos, J.~S. Duncan, B.~van Ginneken,
  A.~Madabhushi, J.~L. Prince, D.~Rueckert, and R.~M. Summers, ``{A review of
  deep learning in medical imaging: Image traits, technology trends, case
  studies with progress highlights, and future promises},'' 2020. [Online].
  Available: \url{http://arxiv.org/abs/2008.09104}
\BIBentrySTDinterwordspacing

\bibitem{Cai2019cycle}
\BIBentryALTinterwordspacing
J.~Cai, Z.~Zhang, L.~Cui, Y.~Zheng, and L.~Yang, ``{Towards cross-modal organ
  translation and segmentation: A cycle- and shape-consistent generative
  adversarial network},'' \emph{Med. Image Anal.}, vol.~52, pp. 174--184, feb
  2019. [Online]. Available:
  \url{https://www.sciencedirect.com/science/article/pii/S1361841518303426}
\BIBentrySTDinterwordspacing

\bibitem{Liu2020JSSR}
\BIBentryALTinterwordspacing
F.~Liu, J.~Cai, Y.~Huo, C.-T. Cheng, A.~Raju, D.~Jin, J.~Xiao, A.~Yuille,
  L.~Lu, C.~Liao, and A.~P. Harrison, ``{JSSR: A Joint Synthesis, Segmentation,
  and Registration System for 3D Multi-Modal Image Alignment of Large-scale
  Pathological CT Scans},'' in \emph{ECCV}, 2020. [Online]. Available:
  \url{http://arxiv.org/abs/2005.12209}
\BIBentrySTDinterwordspacing

\bibitem{Raju2020CoHeter}
\BIBentryALTinterwordspacing
A.~Raju, C.-T. Cheng, Y.~Huo, J.~Cai, J.~Huang, J.~Xiao, L.~Lu, C.~Liao, and
  A.~P. Harrison, ``{Co-Heterogeneous and Adaptive Segmentation from
  Multi-Source and Multi-Phase CT Imaging Data: A Study on Pathological Liver
  and Lesion Segmentation},'' 2020. [Online]. Available:
  \url{http://arxiv.org/abs/2005.13201}
\BIBentrySTDinterwordspacing

\bibitem{Xia2020semi}
\BIBentryALTinterwordspacing
Y.~Xia, D.~Yang, Z.~Yu, F.~Liu, J.~Cai, L.~Yu, Z.~Zhu, D.~Xu, A.~Yuille, and
  H.~Roth, ``{Uncertainty-aware multi-view co-training for semi-supervised
  medical image segmentation and domain adaptation},'' \emph{Med. Image Anal.},
  p. 101766, jun 2020. [Online]. Available:
  \url{https://linkinghub.elsevier.com/retrieve/pii/S1361841520301304}
\BIBentrySTDinterwordspacing

\bibitem{Zhou2019semi}
\BIBentryALTinterwordspacing
Y.~Zhou, Y.~Wang, P.~Tang, S.~Bai, W.~Shen, E.~K. Fishman, and A.~Yuille,
  ``{Semi-supervised 3D abdominal multi-organ segmentation via deep
  multi-planar co-training},'' in \emph{WACV}, 2019, pp. 121--140. [Online].
  Available: \url{https://arxiv.org/pdf/1804.02586.pdf}
\BIBentrySTDinterwordspacing

\bibitem{Liu2020relation}
\BIBentryALTinterwordspacing
Q.~Liu, L.~Yu, L.~Luo, Q.~Dou, and P.~A. Heng, ``{Semi-supervised Medical Image
  Classification with Relation-driven Self-ensembling Model},'' \emph{IEEE
  Trans. Med. Imaging}, 2020. [Online]. Available: \url{https://github.com/
  http://arxiv.org/abs/2005.07377}
\BIBentrySTDinterwordspacing

\bibitem{Wang2020focalmix}
D.~Wang, Y.~Zhang, K.~Zhang, and L.~Wang, ``{FocalMix: Semi-Supervised Learning
  for 3D Medical Image Detection},'' in \emph{CVPR}, 2020.

\bibitem{Huang2017DenseNet}
G.~Huang, Z.~Liu, K.~Q. Weinberger, and L.~van~der Maaten, ``{Densely Connected
  Convolutional Networks},'' in \emph{Proc. IEEE Conf. Comput. Vis. Pattern
  Recognit.}, 2017.

\bibitem{Lin2016Pyramid}
T.~Y. Lin, P.~Doll{\'{a}}r, R.~Girshick, K.~He, B.~Hariharan, and S.~Belongie,
  ``{Feature pyramid networks for object detection},'' in \emph{Proc. IEEE
  Conf. Comput. Vis. Pattern Recognit.}, 2017.

\bibitem{Rebuffi2018domain}
S.~A. Rebuffi, A.~Vedaldi, and H.~Bilen, ``{Efficient Parametrization of
  Multi-domain Deep Neural Networks},'' in \emph{Proc. IEEE Conf. Comput. Vis.
  Pattern Recognit.}, 2018, pp. 8119--8127.

\bibitem{Tian2019FCOS}
Z.~Tian, C.~Shen, H.~Chen, and T.~He, ``{FCOS: Fully Convolutional One-Stage
  Object Detection},'' in \emph{Proc. IEEE Int. Conf. Comput. Vis.}, 2019.

\bibitem{Zhou2019CenterNet}
\BIBentryALTinterwordspacing
X.~Zhou, D.~Wang, and P.~Kr{\"{a}}henb{\"{u}}hl, ``{Objects as Points},'' 2019.
  [Online]. Available: \url{http://arxiv.org/abs/1904.07850}
\BIBentrySTDinterwordspacing

\bibitem{Zhu2019FSAF}
C.~Zhu, Y.~He, and M.~Savvides, ``{Feature Selective Anchor-Free Module for
  Single-Shot Object Detection},'' in \emph{Proc. IEEE Conf. Comput. Vis.
  Pattern Recognit.}, 2019.

\bibitem{Lin2017focal}
T.-Y. Lin, P.~Goyal, R.~Girshick, K.~He, and P.~Doll{\'{a}}r, ``{Focal Loss for
  Dense Object Detection},'' in \emph{Proc. IEEE Int. Conf. Comput. Vis.},
  2017, pp. 2980--2988.

\bibitem{He2016resnet}
K.~He, X.~Zhang, S.~Ren, and J.~Sun, ``{Deep Residual Learning for Image
  Recognition},'' in \emph{Proc. IEEE Conf. Comput. Vis. Pattern Recognit.},
  2016, pp. 770--778.

\bibitem{Yang2019acs}
\BIBentryALTinterwordspacing
J.~Yang, X.~Huang, B.~Ni, J.~Xu, C.~Yang, and G.~Xu, ``{Reinventing 2D
  Convolutions for 3D Medical Images},'' 2019. [Online]. Available:
  \url{http://github.com/m3dv/ACSConv. http://arxiv.org/abs/1911.10477}
\BIBentrySTDinterwordspacing

\bibitem{Yan2019Lesa}
K.~Yan~et al., ``{Holistic and Comprehensive Annotation of Clinically
  Significant Findings on Diverse CT Images : Learning from Radiology Reports
  and Label Ontology},'' in \emph{Proc. IEEE Conf. Comput. Vis. Pattern
  Recognit.}, 2019.

\bibitem{massa2018mrcnn}
F.~Massa and R.~Girshick, ``{maskrcnn-benchmark: Fast, modular reference
  implementation of Instance Segmentation and Object Detection algorithms in
  PyTorch},'' \url{https://github.com/facebookresearch/maskrcnn-benchmark},
  2018, accessed: 03-21-2019.

\bibitem{liu2019radam}
L.~Liu, H.~Jiang, P.~He, W.~Chen, X.~Liu, J.~Gao, and J.~Han, ``On the variance
  of the adaptive learning rate and beyond,'' \emph{arXiv preprint
  arXiv:1908.03265}, 2019.

\bibitem{Carreira2017I3D}
J.~Carreira and A.~Zisserman, ``{Quo Vadis, action recognition? A new model and
  the kinetics dataset},'' in \emph{Proc. IEEE Conf. Comput. Vis. Pattern
  Recognit.}, vol. 2017-Janua, 2017, pp. 4724--4733.

\bibitem{Zhu2020gating}
Z.~Zhu, D.~Jin, K.~Yan, T.-Y. Ho, X.~Ye, D.~Guo, C.-H. Chao, J.~Xiao,
  A.~Yuille, and L.~Lu, ``{Lymph Node Gross Tumor Volume Detection and
  Segmentation via Distance-Based Gating Using 3D CT/PET Imaging in
  Radiotherapy},'' in \emph{MICCAI}, 2020, pp. 753--762.

\bibitem{Zhu2020scatter}
Z.~Zhu, K.~Yan, D.~Jin, J.~Cai, T.~Y. Ho, A.~P. Harrison, D.~Guo, C.~H. Chao,
  X.~Ye, J.~Xiao, A.~Yuille, and L.~Lu, ``{Detecting Scatteredly-Distributed,
  Small, and Critically Important Objects in 3D Oncology Imaging via Decision
  Stratification},'' 2020.

\bibitem{LUNA16Res}
\BIBentryALTinterwordspacing
Grand-challenge.org, ``{LUng Nodule Analysis 2016 Results}.'' [Online].
  Available: \url{https://luna16.grand-challenge.org/Results/}
\BIBentrySTDinterwordspacing

\end{thebibliography}
	
\end{document}